\documentclass{article}
\usepackage{arxiv}
\usepackage{hyperref}
\usepackage{microtype}

\usepackage{amssymb}
\usepackage{amsmath}
\usepackage{amsthm}
\usepackage{mathdots}
\usepackage{mathtools}

\usepackage{tensor}

\usepackage{urwchancal}
\DeclareFontFamily{OT1}{pzc}{}
\DeclareFontShape{OT1}{pzc}{m}{it}{<-> s * [1.10] pzcmi7t}{}
\DeclareMathAlphabet{\mathpzc}{OT1}{pzc}{m}{it}

\usepackage{graphicx}
\usepackage{ifpdf}
\ifpdf
\usepackage{epstopdf}
\PrependGraphicsExtensions{.svg}
\fi
\usepackage{xypic}

\usepackage{graphicx}
\usepackage{algorithmicx,algpseudocode}
\usepackage{algorithm}
\usepackage{amsmath}
\usepackage{amsthm}
\usepackage{amssymb}
\usepackage{tensor}
\usepackage{xfrac}
\usepackage{subcaption}
\usepackage{color}
\usepackage{cancel}
\usepackage{color}
\usepackage{accents}

\newcommand{\idx}[5]{\tensor*[_{\scalebox{0.6}{#5}}^{\scalebox{0.6}{#4}}]{#1}{_{#2}^{\scalebox{0.7}{#3}}}}

\newcommand{\I}{\mathbf{I}}

\newcommand{\rspace}[1]{\mathbb{R}^{#1}}

\newcommand{\ppos}{\mathbf{x}}
\newcommand{\pvel}{\mathbf{v}_\ppos}

\newcommand{\sspace}{S^2}
\newcommand{\ssym}{\eta}
\newcommand{\scoord}{\boldsymbol{\ssym}}

\newcommand{\tsym}{\mu}
\newcommand{\tspace}[1]{T_{\ssym_{#1}}S^2}
\newcommand{\tcoord}{\boldsymbol{\mu}}
\newcommand{\tcoordsize}{\Delta \tsym}

\newcommand{\tmatrix}[1]{\mathbf{P}_{\scalebox{0.6}{$#1$}}}

\newcommand{\betamatrix}[1]{\mathbf{B}_{\scalebox{0.6}{$#1$}}}
\newcommand{\bcoord}{\boldsymbol{\beta}}

\newcommand{\pixcoord}{p}

\newcommand{\vnorm}[1]{\left\|#1\right\|}

\newcommand{\totaldev}[2]{\frac{\mathrm{d} #1}{\mathrm{d} #2}}

\newcommand{\partialdev}[2]{\frac{\partial #1}{\partial #2}}

\newcommand{\dotp}[2]{\langle #1 , #2 \rangle}

\newcommand{\diff}[1]{\delta_{\scalebox{0.5}{$#1$}}}
\newcommand{\diffc}[1]{\delta_{\scalebox{0.5}{$#1^0$}}}
\newcommand{\diffp}[1]{\delta_{\scalebox{0.5}{$#1^+$}}}
\newcommand{\diffm}[1]{\delta_{\scalebox{0.5}{$#1^-$}}}

\newcommand{\img}{Y}
\newcommand{\imgpyr}[3]{\idx{\img}{}{#1}{#2}{#3}}  % Delta hflow
\newcommand{\imghat}{\hat{\img}}
\newcommand{\imggradhat}{\partial_{\scoord} \hat{\img}}                   % estimated image gradient in mu coordinates
\newcommand{\imggradbcoordhat}{\partial_{\bcoord} \hat{\img}}   % estimated image gradient in beta coordinates

\newcommand{\oflow}{\boldsymbol{\Phi}}      % Optical flow
\newcommand{\hflow}{\mathbf{w}}             % Homogeneous flow
\newcommand{\dhflow}{\scalebox{0.6}{$\Delta$} \hflow}         % Delta H-flow
\newcommand{\hflowpyr}[3]{\idx{\hflow}{}{#1}{#2}{#3}}  % hflow pyramid
\newcommand{\dhflowpyr}[3]{\idx{\dhflow}{}{#1}{#2}{#3}}  % Delta hflow pyramid

\newcommand{\depth}{\lambda}                % depth

\newcommand{\invdepth}{\rho}                % inverse depth
\newcommand{\invdepthhat}{\hat{\invdepth}}
\newcommand{\invdepthgradhat}{\partial_{\scoord} \invdepthhat}
\newcommand{\invDepthBcoordGradHat}[1]{\partial_{\bcoord_{#1}} \invdepthhat}
\newcommand{\invdepthpyr}[3]{\idx{\invdepth}{}{#1}{#2}{#3}}  % inverse depth pyramid
\newcommand{\invdepthpyrhat}[3]{\idx{\hat{\invdepth}}{}{#1}{#2}{#3}}  % inverse depth pyramid

\newcommand{\camvel}{\mathbf{v}_c}
\newcommand{\camacc}{\boldsymbol{a}_c}
\newcommand{\camangvel}{\Omega}

\newcommand{\vel}{\mathbf{v}}

\newcommand{\acc}{\boldsymbol{a}}

\usepackage{xcolor}  % Needed for colours - load before mdframed

\usepackage{mdframed} %% useful for shading large blocks of text
\usepackage{lipsum}  % Needed to implment mdframed.

\definecolor{NiceGreen}{HTML}{F5FFEE}

\newmdenv[%
leftmargin = 0,%
rightmargin = 0,%
backgroundcolor = NiceGreen,%
innertopmargin = \topskip,%
splittopskip = \topskip,%
skipabove =\baselineskip,%
skipbelow = \baselineskip]
{RMnotes}
\usepackage{comment}
\specialcomment{notes}{\begin{RMnotes}}{\end{RMnotes}}

\newcommand{\papercitation}{% add citation to published paper in text
Pre-print
}

\newcommand{\publicationversion}
{Pre-print}
\newcommand{\publicationdetails}
{
\papercitation 
}
\begin{document}
%===============================================================================

\title{Real-time Structure Flow}
% \headertitle{SHORT TITLE}
%% The headertitle will be printed in the header.  If the main title is too long define a shorter headertitle

\author{
\href{https://orcid.org/0000-0003-2203-3535}{\includegraphics[scale=0.06]{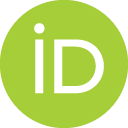}\hspace{1mm}
Juan David Adarve}
\\
	Systems Theory and Robotics Group \\
	Australian National University \\
    ACT, 2601, Australia \\
	\texttt{juan.adarve@alumni.anu.edu.au} \\
	\And	\href{https://orcid.org/0000-0002-7803-2868}{\includegraphics[scale=0.06]{orcid.png}\hspace{1mm}
Robert Mahony}
\\
	Systems Theory and Robotics Group \\
	Australian National University \\
    ACT, 2601, Australia \\
	\texttt{robert.mahony@anu.edu.au} \\
}

\maketitle

%===============================================================================
\begin{abstract}
	This article introduces the \emph{structure flow} field; a flow field that can provide high-speed robo-centric motion information for motion control of highly dynamic robotic devices and autonomous vehicles.
	Structure flow is the \emph{angular 3D velocity} of the scene at a given pixel.
	We show that structure flow posses an elegant evolution model in the form of a Partial Differential Equation (PDE) that enables us to create dense flow predictions forward in time.
	We exploit this structure to design a predictor-update algorithm to compute structure flow in real time using image and depth measurements.
	The prediction stage takes the previous estimate of the structure flow and propagates it forward in time using a numerical implementation of the structure flow PDE.
	The predicted flow is then updated using new image and depth data.
	The algorithm runs up to 600 Hz on a Desktop GPU machine for 512x512 images with flow values up to 8 pixels.
	We provide ground truth validation on high-speed synthetic image sequences as well as results on real-life video on driving scenarios.
\end{abstract}

% \keywords{
% XXXX
% }

%===============================================================================
%%%%%%%%%%%%%%%%%%%%%%%%%%%%%%%%%%%%%%%%%%%%%%%%%%%%%%%%%%%
% INTRODUCTION
%%%%%%%%%%%%%%%%%%%%%%%%%%%%%%%%%%%%%%%%%%%%%%%%%%%%%%%%%%%
\section{Introduction}
\label{sec:introduction}

Reliable and robust motion control is a central component of all autonomous robotic systems.
As robotic vehicles become more mobile and begin to operate in highly unstructured and dynamic environments, such as urban streetscapes, the requirements on the perception subsystem of the robot are correspondingly more demanding.
%, providing the situational awareness fundamental for obstacle avoidance and motion control,.
Classically, most autonomous robot use a combination of Simultaneous Localisation And Mapping (SLAM) \cite{2016_Cadena} along with dedicated sensor-control feedback for motion control.
More specifically, SLAM algorithms are used to provide large scale (sparse) environment representations appropriate for path planning and decision processes, while real-time motion control loops are closed using high-bandwidth direct sensor feedback.
Sensor systems capable of providing a dense, high-bandwidth, 3D spatiotemporal (structure and motion) representation of an outdoor environment, such as LIDAR, are expensive, and often bulky and heavy.
In contrast, high frame-rate vision systems offer a rich, dense, and dynamic sensor modality for low-weight and low-price.
Vision systems, however, require sophisticated post processing to extract spatiotemporal information.
Modern dense visual SLAM implementations \cite{2015_Meilland,2016_Whelan} demonstrate the potential of vision systems to provide situational awareness to autonomous robotic vehicles, although these algorithms still suffer from limitations such as; (relatively) low update rates, static environment assumptions and limited map size.

We propose the following aspirational goal requirements for a modern vision based spatiotemporal perception system capable of providing real-time situational awareness for a robot moving in a highly unstructured and dynamic environment such as a streetscape:
\begin{itemize}
    \item Egocentric (robo-centric) map representation.

    \item Dense range estimation to the environment. \emph{Benchmark: angular resolution 0.1 degrees.}

    \item Dense 3D-velocity (structure flow) estimation.  \emph{Benchmark: resolution 0.05rad.s$^{-1}$, (for example, 0.05m.s$^{-1}$ error in 1m.s$^{-1}$ velocity at 1m distance).}

    \item Real-time (high-bandwdith) updates.   \emph{Benchmark: updates at 300 Hz.}
\end{itemize}
Such a system would provide sufficient information for robust local path planning and obstacle avoidance.
High level path planning and decision processes could be undertaken separately based on more traditional large scale SLAM algorithms.
The present work also continues development (see also \cite{2016_Adarve}) of a new paradigm of vision processing algorithms based on real-time filtering concepts, that we believe will provide an important tool in the future development of autonomous robotic systems.

The technical contributions of this article are twofold.
First, we introduce a new robo-centric spatiotemporal representation of motion, the \emph{structure flow}.
The structure flow is a 3-vector assigned to each `pixel' in the image comprised of the three-dimensional Euclidean velocity between the robot and the environment (the scene flow \cite{1999_Vedula}) scaled by the inverse range of the scene.
Analogously to optic flow, structure flow encodes information about the visually perceived motion and ego-motion of the scene.
However, structure flow generalises optic flow by including a component in the normal direction of the image that is closely linked to visual divergence.
Structure flow will have a similar utility as optic flow for control of robotic vehicles \cite{2002_Hamel,2005_Ruffier,2008_Herisse,2011_Srinivasan} with the added advantage that the normal component of structure flow directly yields the divergent flow properties used in many of the obstacle avoidance algorithms \cite{2012_McCarthy}.
A key advantage of structure flow is that, for reasonable assumptions on the environment and robot motion, it is possible to derive a dynamic model of the field evolution in the form of a Partial Differential Equation (PDE).
Moreover, the structure flow PDE follows the same pattern as the partial differential equation modeling the evolution of the inverse depth field \cite{2009_Bonnabel,2012_Zarrouati}.
We exploit this to forward propagate estimates of the structure flow and the inverse depth in image coordinates to make predictions of future motion and structure of the environment.

The second technical contribution of the paper lies in proposing an iterative algorithm to compute structure flow in real-time from stereo image data.
The approach taken is a highly distributed predictor-update algorithm that we implement on a GPU.
The prediction step is a numerical implementation of a PDE integration
scheme that propagates the current estimate of the structure flow forward in time.
Error in the structure flow estimate can then be estimated across a sequence of temporal frames, and we use a simple least-squares regularised update to iteratively correct errors in the structure flow estimate.
The resulting algorithm runs on a Nvidia GTX-780 GPU, processing $512 \times 512$ at approximately 600 Hz for flow vectors up to 8 pixels in magnitude.
The relative performance of the algorithm far outperforms (10-20 times faster) all scene flow algorithms in the literature in terms of processing speed.
Although it is less accurate than some classical algorithms, its advantages in speed, dense estimation, and robustness make it highly suitable for real-world mobile robotic applications.

%%%%%%%%%%%%%%%%%%%%%%%%%%%%%%%%%%%%%%%%%%%%%%%%%%%%%%%%%%%
% RELATED WORK
%%%%%%%%%%%%%%%%%%%%%%%%%%%%%%%%%%%%%%%%%%%%%%%%%%%%%%%%%%%
\section{Related Work}
\label{sec:related_work}

The estimation of motion from image sequences is a heavily studied problem in computer vision and robotics \cite{1994_Barron,2011_Baker}.
The importance of these motion fields in the control of mobile robotic vehicles is equally important, and has been studied for more than thirty years in applications such as visual control of aerial vehicles \cite{2005_Ruffier,2011_Srinivasan,2012_Herisse} and visual odometry \cite{2011_Scaramuzza}.

The most common motion field considered is the \emph{optical flow} field, defined as the displacement of points in the image plane \cite{1994_Barron}.
Algorithms for the computation of flow can be classified as local or global based methods, from which the seminal works of Lucas and Kanade (local) \cite{1981_Lucas} and Horn and Schunck \cite{1981_Horn} (global) marked a distinction between local and global methods. Recent algorithms can be found in benchmark datasets such as Middlebury \cite{2011_Baker} and Kitti \cite{2013_Geiger}.
In robotics, an underlying requirement for flow algorithms is their capability to run at real-time frequencies.
Real time methods \cite{2014_Plyer,2016_Adarve} are usually local-based methods, which can be efficiently implemented using GPU or FPGA hardware.
The vast majority of optical flow algorithms developed in by the computer-vision community in the last ten years are focused on accuracy and not on computational efficiency \cite{2011_Baker,2013_Geiger}.

A more recent visual flow is \emph{scene flow}, defined as the three-dimensional velocity field of points moving in the world \cite{1999_Vedula} expressed in the `image' frame of the robot.
Stereoscopic algorithms use image sequences captured using a stereo camera to estimate the 3D scene flow.
Early work by Patras \textit{et~al.} \cite{1996_Patras} formulates the joint estimation of disparity and motion (scene flow) from stereo sequences.
Huguet and Devernay \cite{2007_Huguet} proposed a variational approach for the joint computation of disparity and optical flow. This formulation of optical flow plus temporal disparity change is known as \emph{disparity flow} \cite{2006_Gong}.
Wedel \textit{et~al.} \cite{2011_Wedel} decouple the variational formulation of Huguet \textit{et.~al.}  to compute flow and disparity in two separate stages while preserving stereo constraints.
Vogel \textit{et.~al.} \cite{2015_Vogel} compute scene flow by means of a piece-wise planar segmentation of stereo images for which a 3D rotation and translation (scene flow) is computed.
The accuracy of recent two-pair stereo based algorithms is compiled on the Kitti scene flow dataset by Menze and Geiger \cite{2015_Menze}.
Algorithms using RGB-D sensors to compute scene flow combine monocular images and depth measurements to estimate the 3D velocity field.
Letouzey \textit{et.~al.}  \cite{2011_Letouzey} combine sparse features points and dense smoothness constraints to estimate dense scene flow from RGB-D data. Hadfield and Bowden \cite{2011_Hadfield} use a particle filter formulation to track points in the scene.
Quiroga \textit{et.~al.} \cite{2012_Quiroga} use brightness and depth data in a Lukas-Kanade type tracking framework to compute scene flow parametrized as a rotation plus translation transform.

If only depth/range measurements are available it is still possible to compute \emph{range flow}.
Yamamoto  \textit{et.~al.} derived the range flow constraint \cite{1993_Yamamoto}, a partial differential equation to model the change of depth.
Differential methods to compute range flow suffer from a 3D version of the aperture problem similar to that found in optical flow methods \cite{2002_Spies}.
Herbst \textit{et.~al.} \cite{2013_Herbst} use this differential constraint on depth and combined with color image to derive a variational method to compute scene flow following a formulation similar to Brox \textit{et.~al.} \cite{2004_Brox} for computing optical flow.

In terms of real-time capable dense 3D motion estimation algorithms, most methods use Graphics Processing Units (GPU) for implementing per-pixel level operations in parallel.
Gong \cite{2009_Gong} reports 12 Hz optical and disparity flow estimation on QVGA image resolution ($320 \times 240$).
Rabe \textit{et.~al.} \cite{2010_Rabe} reported 25 Hz frequency on VGA ($640 \times 480$) image sequences using GPU.
The algorithm of Wedel \textit{et.~al.} \cite{2011_Wedel} reports frame rates of 20 Hz also at QVGA resolution.
RGB-D flow by Herbst \textit{et.~al.} \cite{2013_Herbst} runs between 8-30 Hz at QVGA depending on the amount of smoothing required.

%%%%%%%%%%%%%%%%%%%%%%%%%%%%%%%%%%%%%%%%%%%%%%%%%%%%%%%%%%%
% PROBLEM FORMULATION
%%%%%%%%%%%%%%%%%%%%%%%%%%%%%%%%%%%%%%%%%%%%%%%%%%%%%%%%%%%
\section{Structure Flow}
\label{sec:structure_flow}

We use a spherical camera model with the embedded coordinates $\sspace = \{ \scoord \in \rspace{3} \;|\; \vnorm{\scoord} = 1\}$ representing points on the sphere.
The tangent space $\tspace{}$ associated to each point $\scoord$ is defined as
\begin{equation}
  \tspace{} = \{ \tcoord \in \rspace{3} \;|\; \dotp{\scoord}{\tcoord} =0 \}
\end{equation}
and the projection matrix $\tmatrix{\scoord} : \rspace{3} \rightarrow \tspace{}$ is the orthogonal projection onto $\tspace{}$
\begin{equation}
    \label{eq:tmatrix}
    \tmatrix{\scoord} = (\I_{3} - \scoord \scoord^\top).
\end{equation}

\subsection{Optical, scene and structure flow}

\begin{figure}[h]
    \centering
    \includegraphics[width=0.35\textwidth]{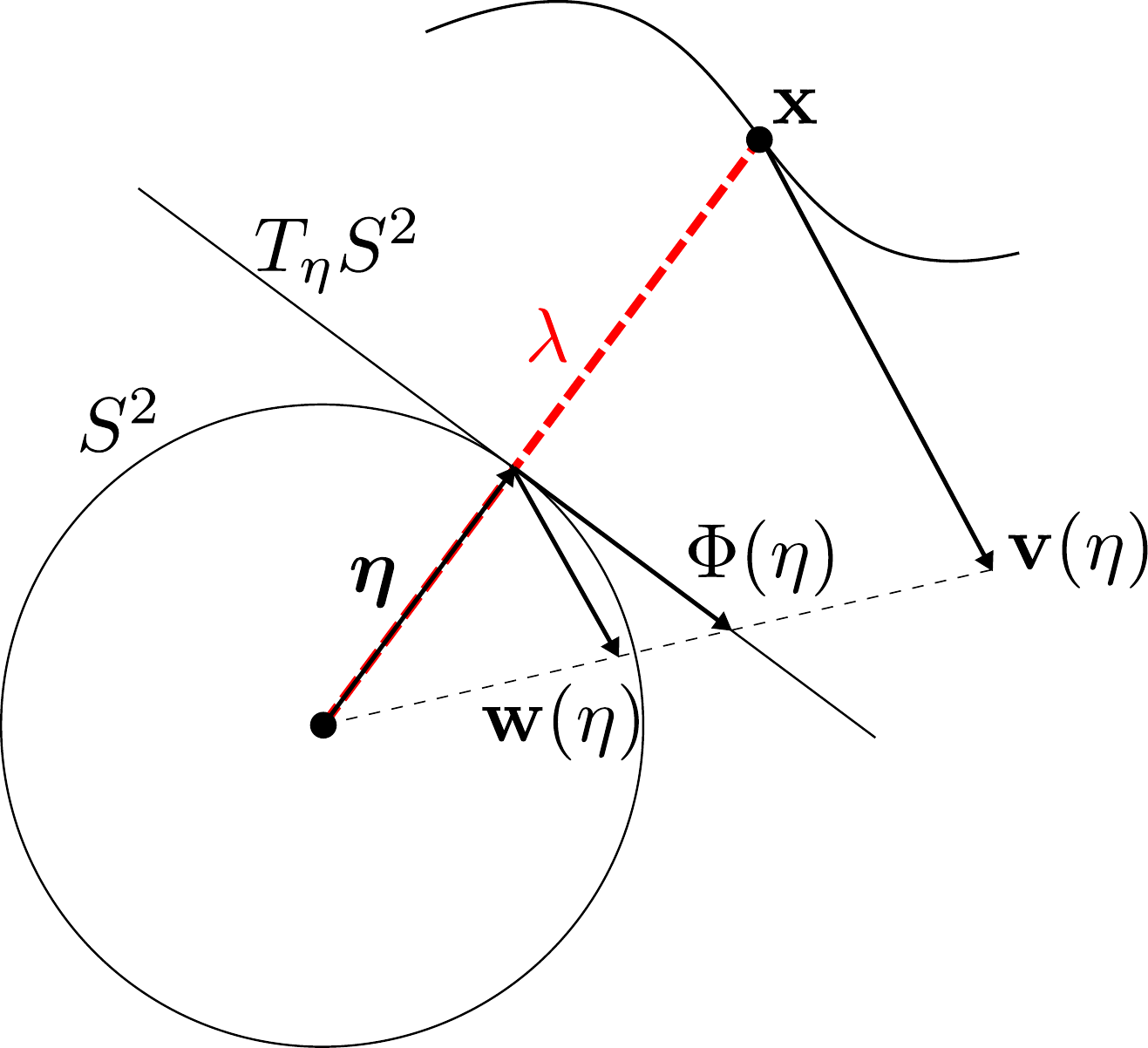}
    \caption{Scene $\vel(\scoord)$, structure $\hflow(\scoord)$ and optical flow $\oflow(\scoord)$.}
    \label{fig:hflow}
\end{figure}

Consider Figure \ref{fig:hflow} where the camera is located in the centre of the body-fixed frame.
Let $\camvel$ and $\camacc$ denote the linear velocity and acceleration respectively of the camera, taken with respect to an inertial frame, and expressed in the camera frame.
Let $\camangvel$ denote the angular velocity of the camera expressed in the camera frame and let $\camangvel_\times$ be the skew-symmetric matrix such that $\Omega \times v = \Omega_\times v$ for any vector $v$.
For the present derivation we will assume that $\Omega$ and $\camacc$ are approximately constant.
This is a reasonable assumption given the very high frame rates of the algorithms that we propose.
With this assumption, the second order kinematics of the camera are
\begin{align}
    \totaldev{\camvel}{t} &= -\camangvel_\times \camvel + \camacc \label{eq:camvel_dot}\\
    \totaldev{\camangvel}{t} &= 0 \label{eq:totaldev_camangvel}
\end{align}
Both $\camangvel$ and $\camacc$ are readily available from a typical inertial measurement unit and we assume they are known.

Let $\ppos = \ppos(t) \in \rspace{3}$ denote the position of a point in the scene at time $t$ in body fixed frame (Figure~\ref{fig:hflow}).
Let $\pvel$ denote the velocity of the point $\ppos$ with respect to the inertial frame and expressed in the camera body fixed frame.
We assume that the scene is moving with a constant velocity with respect to the inertial frame.
With these assumptions, the first and second order kinematics of a point $\ppos$ in the scene expressed in the camera frame are:
\begin{align}
    \totaldev{\ppos}{t} &= - \camangvel_\times \ppos + \pvel - \camvel , \label{eq:totaldev_ppos} \\
    \totaldev{\pvel}{t} &= -\camangvel_\times \pvel.     \label{eq:totaldev_pvel}
\end{align}

The approach taken in this paper is to parameterize the environment surrounding the robot by direction $\scoord \in \sspace$.
Thus, at time $t$, the first visible point in the environment in direction $\scoord$ is denoted $\ppos(\scoord, t)$.
The range or depth map around the robot is a single valued depth field $\depth: \sspace \times \rspace{}  \rightarrow \rspace{}$,
\begin{equation}
    \label{eq:depth}
    \depth(\scoord, t) = \dotp{\ppos(\scoord,t)}{\ppos(\scoord,t)}^{\sfrac{1}{2}}.
\end{equation}
Although range $\depth(\scoord,t)$ appears to be the more intuitive variable, it turns out that it is more natural in the context of the mathematical development to use an inverse ratio of range $\invdepth : \sspace \times \rspace{} \rightarrow \rspace{}$,
\begin{equation}
    \label{eq:inv_depth}
    \invdepth(\scoord, t) = \frac{\depth_\text{ref}}{\lambda(\scoord, t)},
\end{equation}
where $\depth_\text{ref}$ is the reference range to the spherical image plane.
We will choose $\depth_\text{ref} = 1$m for simplicity and suppress the $\depth_\text{ref}$ notation in the remainder of the paper. 
However, one should still think of $\invdepth(\scoord, t)$ as a dimensionless ratio of ranges related to the perceived visual angle of an object at range $\depth(\scoord, t)$ as observed on the sphere of radius $\depth_\text{ref}$. In particular, $\invdepth(\scoord, t)$ is intrinsically a visual measure or angle, whereas $\depth(\scoord, t)$ is extrinsically related to the physical scene and has units of metres.

The scene flow is defined as the 3D velocity of the environment relative to the camera \cite{1999_Vedula} of a point $\ppos(\scoord,t)$ in the scene.
To compute scene flow, consider a solution $\ppos(\tau)$ of \eqref{eq:totaldev_ppos} for $\tau \in (t - \delta, t + \delta)$, $\delta > 0$, in a small time interval around an `initial condition' $\ppos(t) = \ppos(\scoord, t)$ in the desired direction $\scoord$ at time $t$.
Choose $\scoord(\tau)$ to evolve such that $\ppos(\tau) = \ppos(\scoord(\tau),\tau)$ is the point in the scene associated with the evolution of $\ppos(\tau)$ and note that $\scoord(t) = \scoord$; that is $\scoord(\tau)$ evaluated at $\tau = t$ is $\scoord$.
The scene flow is the vector field $\vel : \sspace \times \rspace{} \rightarrow \rspace{3}$ defined as
\begin{equation}
    \vel(\scoord, t) = \left. \totaldev{\ppos(\scoord(\tau), \tau)}{\tau}\right|_{\tau = t}  = - \camangvel_\times \ppos + \pvel - \camvel. \label{eq:scene_flow}
\end{equation}
It is important to note that, whereas \eqref{eq:totaldev_ppos} tracks a particle in the environment, the scene-flow is a field that assigns values for all points parameterized on the sphere.
In particular, $\scoord$ and $t$ are independent variables in the expression \eqref{eq:scene_flow} for $\vel(\scoord, t)$.

A more common notation in the literature used for the total derivative is
\[
\totaldev{\ppos}{t}(t) =
\left. \totaldev{\ppos(\scoord(\tau), \tau)}{\tau}\right|_{\tau = t}
\]
where the underlying construction of the segment of trajectory $\ppos(\tau)$ is understood.
We have chosen to use a more explicit notation in this paper to provide more clarity in the derivations.

We define the structure flow field $\hflow : \sspace \times \rspace{} \rightarrow \rspace{3}$ as the three-dimensional vector field consisting of the scene flow scaled by the inverse depth
\begin{align}
    \hflow(\scoord, t) &= \frac{1}{\depth(\scoord,t)} \vel(\scoord, t) \nonumber \\
        &= - \Omega_\times \scoord + \frac{1}{\depth(\scoord,t)} (\pvel - \camvel)  \label{eq:homogeneous_flow}
\end{align}
Note that while scene flow has units of m.s$^{-1}$, structure flow has units of rad.s$^{-1}$, where the angle is related to the ratio of distances, as would be expected of an image based flow measure.
We also introduce notation to decompose structure flow into rotational and stabilised (linear) components, $\hflow = \hflow_r + \hflow_s $, where
\begin{align}
    \hflow_r(\scoord,t) &= - \Omega_\times \scoord, \label{eq_hflow_r} \\
    \hflow_s(\scoord,t) &= \frac{1}{\depth(\scoord,t)} (\pvel - \camvel). \label{eq:hflow_s}
\end{align}

In order to compute evolution equations for structure flow it is necessary to compute the total time derivative of the flow.
We first compute total time derivatives of $\scoord(\tau)$ and $\depth(\tau)$ defined along trajectories $\ppos(\tau) = \ppos(\scoord(\tau),\tau)$ induced by the equations of motion \eqref{eq:camvel_dot},  \eqref{eq:totaldev_camangvel}, \eqref{eq:totaldev_ppos} and \eqref{eq:totaldev_pvel}.
For depth $\depth(\tau)$, one has
\begin{align}
  \left.  \totaldev{\depth(\tau)}{\tau} \right|_{\tau = t}
  &=  \left. \totaldev{}{\tau} \dotp{\ppos(\tau)}{\ppos(\tau)}^{\sfrac{1}{2}} \right|_{\tau = t} \notag \\
  & = \left. \frac{\ppos(\tau)^\top}{\dotp{\ppos(\tau)}{\ppos(\tau)}^{\sfrac{1}{2}}} \totaldev{\ppos(\tau)}{\tau}  \right|_{\tau = t}       \notag \\
    &=  \left. \dotp{\scoord(\tau)}{\totaldev{\ppos(\tau)}{\tau}}\right|_{\tau = t}  \notag \\
    & = \depth(\scoord,t) \dotp{\scoord}{\hflow(\scoord,t)}, \label{eq:totaldev_depth}
\end{align}
where the last line follows by substituting for structure flow \eqref{eq:homogeneous_flow} at $\tau = t$.
A similar derivation for the inverse depth $\invdepth(t)$ yields
\begin{align}
   \left.  \totaldev{\invdepth(\tau)}{\tau} \right|_{\tau = t}
& = -\invdepth(\scoord,t) \dotp{\scoord}{\hflow(\scoord,t)} \label{eq:totaldev_invdepth}
\end{align}
We note that the right hand side of both \eqref{eq:totaldev_depth} and \eqref{eq:totaldev_invdepth} depend only on the field variables $\hflow$, $\depth$ and $\invdepth$ defined at $(\scoord,t)$ and not on the trajectory $\ppos(\tau)$ from which they are derived.

Next, consider the rate of change of the direction $\scoord(\tau) = \sfrac{\ppos(\tau)}{\depth(\tau)}$.
One has
\begin{align}
  \left.  \totaldev{\scoord(\tau)}{\tau} \right|_{\tau = t}
  &= \left. \totaldev{}{\tau} \left( \frac{\ppos(\tau)}{\depth(\tau)} \right) \right|_{\tau = t} \\
  & = \left.  \frac{\totaldev{\ppos(\tau)}{\tau} \depth(\tau) - \ppos(\tau) \totaldev{\depth(\tau)}{\tau} }{\depth(\tau)^2} \right|_{\tau = t} \nonumber  \\
    &= \left. \frac{1}{\depth(\tau)} \left( \totaldev{\ppos(\tau)}{\tau} - \scoord(\tau) \dotp{\scoord(\tau)}{\totaldev{\ppos(\tau)}{\tau}} \right)  \right|_{\tau = t}  \nonumber \\
    &= \frac{1}{\depth(\scoord,t)} (\I - \scoord \scoord^\top) \vel(\scoord,t)   \nonumber \\
    &= \tmatrix{\scoord} \hflow(\scoord,t) \label{eq:optical_flow}
\end{align}
Note that $\tmatrix{\scoord}\hflow$ satisfies $\dotp{\scoord}{\tmatrix{\scoord}\hflow} = 0$ and hence, is an element of the tangent space of $\scoord$.
Indeed, the total time derivative of $\scoord(\tau)$ is the optical flow $\oflow(\scoord, t) = \tmatrix{\scoord}\hflow$ perceived by the spherical camera \cite{1987_Koenderink,2002_Hamel}.

Finally, consider the total derivative of the structure flow vector $\hflow(\tau) = \hflow(\scoord(\tau),\tau)$ taken along a trajectory induced by the equations of motion \eqref{eq:camvel_dot},  \eqref{eq:totaldev_camangvel}, \eqref{eq:totaldev_ppos} and \eqref{eq:totaldev_pvel}.
One has
\begin{align}
   \left.  \totaldev{\hflow(\tau)}{\tau}  \right|_{\tau = t}
   & = \left. \totaldev{}{\tau} \left(- \camangvel_\times \scoord(\tau) + \frac{1}{\depth(\tau)} (\pvel(\tau) - \camvel(\tau) ) \right) \right|_{\tau = t}  \notag \\
    &=
- \left.  \totaldev{}{\tau} \left( \camangvel_\times \scoord(\tau) \right) \right|_{\tau = t} +
    \left. \totaldev{\rho(\tau)}{\tau} (\pvel - \camvel) \right|_{\tau = t} \notag \\
& \quad\quad\quad\quad\quad +   \rho(\tau) \left. \totaldev{}{\tau} (\pvel(\tau) - \camvel(\tau))  \right|_{\tau = t}
 \notag \\
    &= - \camangvel_\times \tmatrix{\scoord} \hflow(\scoord,t) \notag\\
    & \quad\quad\quad
    - \rho(\scoord) \dotp{\scoord}{\hflow(\scoord,t)} (\pvel(t) - \camvel(t)) \notag \\
    & \quad\quad\quad\quad  - \rho(\scoord) \camangvel_\times (\pvel(t) - \camvel(t))  + \invdepth(\scoord) \camacc
    \notag \\
    &= - \camangvel_\times \tmatrix{\scoord} \hflow(\scoord,t)
    -  \hflow_s(\scoord,t) \dotp{\scoord}{\hflow(\scoord,t)} \notag \\
    & \quad\quad\quad
    - \camangvel_\times \hflow_s(\scoord,t)  + \invdepth(\scoord) \camacc
    \notag
\end{align}
We define a grouped acceleration term
\begin{align*}
    \acc_\hflow (\scoord,t) & = \invdepth \camacc  - \camangvel_\times \hflow_s(\scoord,t)
& = \invdepth \camacc  - \camangvel_\times \left( \hflow  + \camangvel_\times \scoord \right).
\end{align*}
This is the exogenous linear acceleration of the structure flow field at $(\scoord,t)$ due to camera motion along with the coriolis term $\camangvel_\times \hflow_s(\scoord,t)$ associated with the fact that the linear velocities are expressed in a rotating frame.
Recall that  $\hflow_s = \hflow - \hflow_r$ and $\hflow_r = -\Omega_\times \scoord$.
Rewriting in terms of the full structure flow and $\acc_\hflow$ one has
\begin{align}
   \left.  \totaldev{\hflow(\tau)}{\tau}  \right|_{\tau = t}
=&- \camangvel_\times \tmatrix{\scoord} \hflow(\scoord,t) + \Omega_\times \scoord  \dotp{\scoord}{\hflow(\scoord,t)} \notag \\
 &\quad\quad\quad\quad -\hflow(\scoord,t) \dotp{\scoord}{\hflow(\scoord,t)} + \acc_\hflow(\scoord,t) \notag \\
    =& -\hflow(\scoord,t) \dotp{\scoord}{\hflow(\scoord,t)} -\camangvel_\times \hflow(\scoord,t) + \acc_\hflow(\scoord,t).\label{eq:totaldev_hflow}
\end{align}
Here, the term $- \camangvel_\times \hflow$ is to be expected since $\hflow$ is expressed with respect to the rotating camera frame.
The driving term $-\hflow \dotp{\scoord}{\hflow}$ is the autonomous growth or decrease in $\hflow$ due to its range velocity component and the field acceleration $\acc_\hflow$ was defined earlier.

\begin{figure}[!t]
    \centering
    \includegraphics[width=0.6\textwidth]{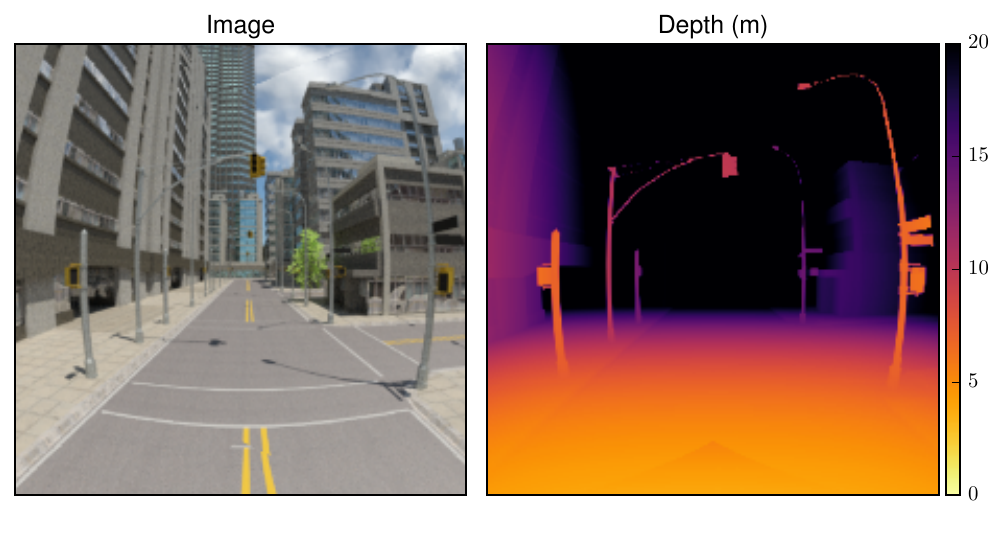} \\
    \includegraphics[width=0.6\textwidth]{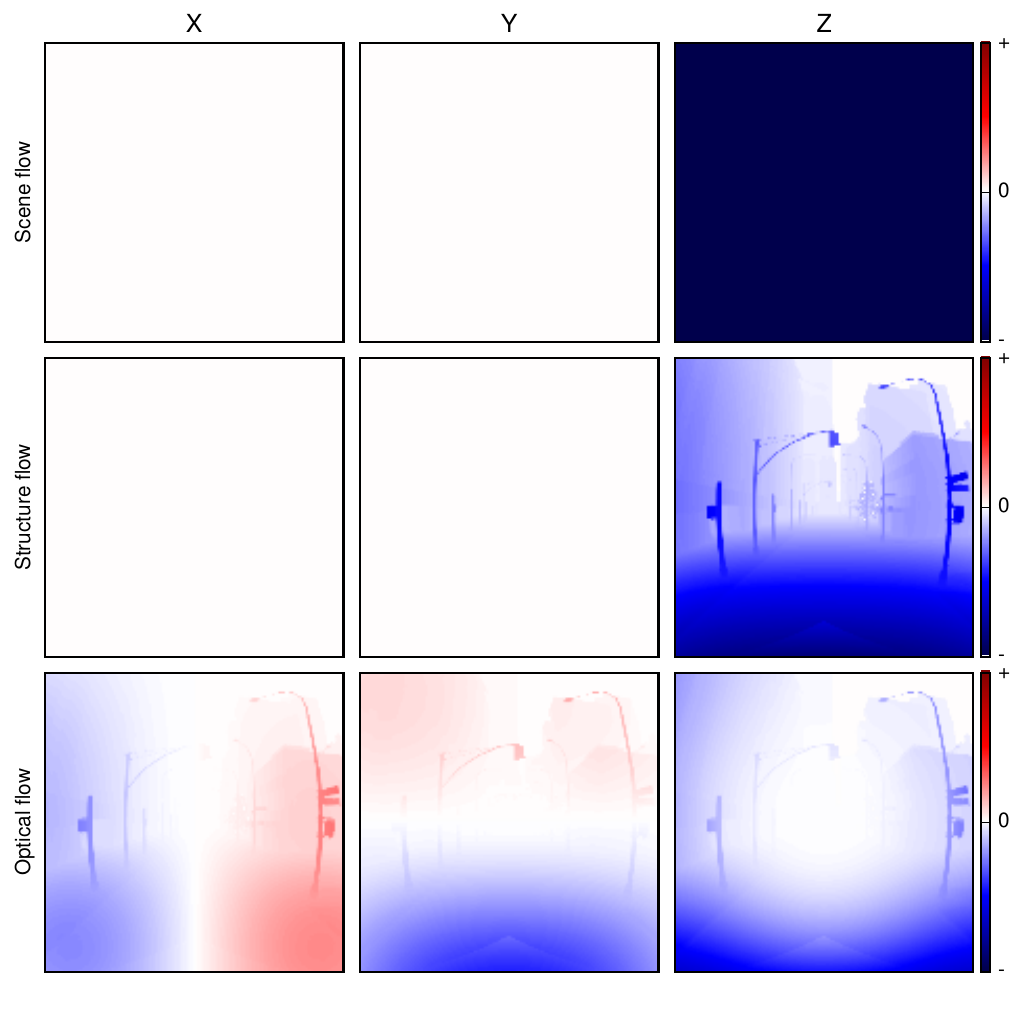}
    \caption{Scene, structure and optical flow fields for a forward moving camera in a static scene.}
    \label{fig:flow_comparison}
\end{figure}

Figure \ref{fig:flow_comparison} illustrates the $xyz$ components of scene, structure and optical flow fields for a static scene and a forward moving camera (in the $z$ direction) with no rotation. Since the scene is static, the scene flow equals $-\camvel$ for all pixels  and in particular, the $xy$ components of scene flow are zero.
The structure flow combines the scene flow and the inverse depth at each pixel to create a velocity field that differentiates between close and distant objects. In contrast, the optical flow generates a divergent velocity field product of the projection onto tangent space. The focus of expansion of the optical flow at the image center corresponds to the direction of motion.

%%%%%%%%%%%%%%%%%%%%%%%%%%%%%%%%%%%%%%%%%%%%%%%%%%%%%%%%%%%
% CONSERVATION EQUATIONS
%%%%%%%%%%%%%%%%%%%%%%%%%%%%%%%%%%%%%%%%%%%%%%%%%%%%%%%%%%%
\subsection{Evolution equations}
\label{section:conservation_equations}

This section derives the fundamental partial differential equations that model the temporal change of image brightness, structure flow and inverse depth on the spherical camera.

%%%%%%%%%%%%%%%%%%%%%%%%%%%%%
% IMAGE BRIGHTNESS
%%%%%%%%%%%%%%%%%%%%%%%%%%%%%
\subsubsection{Image brightness}

Let $\img(\scoord, t)$ denote the image brightness in the direction $\scoord \in \sspace$ at time $t$ associated to point $\ppos(t)$ in the scene.
That is $\img : \sspace \times \rspace{} \rightarrow \rspace{}$ is modeled as a scalar field on the sphere.
A common assumption on differential methods for computing either optical or scene flow is that image brightness is constant and the surface is Lambertian.
That is, the brightness value corresponding to point $\ppos(t)$ as seen by the camera does not change with the evolution of $\ppos(t)$ in time.
This assumption corresponds to the well known constant brightness condition on the \emph{total derivative} of the image brightness field,
\begin{equation}
    \label{eq:image_total_derivative}
    \totaldev{\img}{t}(t) = \left. \totaldev{Y(\scoord(\tau))}{\tau} \right|_{\tau = t}= 0.
\end{equation}

Expressing the total derivative of $\img(\scoord, t)$ in terms of its partial derivatives one obtains

\begin{equation}
    \label{eq:image_conservation_pde}
    \partialdev{\img}{\scoord} \totaldev{\scoord}{t} + \partialdev{\img}{t} = \totaldev{\img}{t}
\end{equation}
Here $\partialdev{\img}{\scoord} \in \tspace{}$ denotes the image gradient and it is an element of the tangent space of $\scoord$.
Moreover, $\totaldev{\scoord}{t} = \left. \totaldev{\scoord(\tau)}{\tau} \right|_{\tau = t} = \oflow(\scoord, t)$ is the optical flow vector at $\scoord$ defined in Equation \eqref{eq:optical_flow}.
Expressing PDE \eqref{eq:image_conservation_pde} in terms of structure flow, one has
\begin{equation}
    \label{eq:image_conservation}
    \partialdev{\img}{t} = -\partialdev{\img}{\scoord} \tmatrix{\scoord}\hflow
\end{equation}
Equation \eqref{eq:image_conservation} is the classical brightness conservation equation \cite{1994_Barron} expressed in spherical camera coordinates and with respect to structure flow.

%%%%%%%%%%%%%%%%%%%%%%%%%%%%%
% structure flow
%%%%%%%%%%%%%%%%%%%%%%%%%%%%%
\subsubsection{Structure flow}

Analogous to the brightness conservation, we consider $\hflow(\scoord, t)$ as a continuous vector field on the sphere. The relative change with respect to $\scoord$ and $t$ is

\begin{equation}
    \label{eq:hflow_partialdevs}
    \partialdev{\hflow}{\scoord}\totaldev{\scoord}{t} + \partialdev{\hflow}{t} = \totaldev{\hflow}{t}
\end{equation}
where $\partialdev{\hflow}{\scoord}$ is the Jacobian matrix of $\hflow$ at $\scoord$ and $\totaldev{\hflow}{t} = \left. \totaldev{\hflow(\scoord(\tau),\tau)}{\tau} \right|_{\tau = t}$ is the optical flow as derived in \eqref{eq:optical_flow}.
In this case, the total flow is not conserved, that is, $\totaldev{\hflow}{t} \neq 0$, and one must substitute \eqref{eq:totaldev_hflow} to derive the associated PDE
\begin{equation}
    \label{eq:hflow_conservation}
    \partialdev{\hflow}{t} = -\partialdev{\hflow}{\scoord}\tmatrix{\scoord} \hflow -\hflow \dotp{\scoord}{\hflow} - \camangvel_\times \hflow + \acc_\hflow
\end{equation}
Equation \eqref{eq:hflow_conservation} describes a non-linear transport process where the structure flow is propagated on the sphere at a velocity equal to the induced optical flow $\tmatrix{\scoord} \hflow$.
Additionally, the source terms inject or remove energy from the structure flow field according to the kinematics of the robot and the scene relative to it.

%%%%%%%%%%%%%%%%%%%%%%%%%%%%%
% Depth and inverse depth
%%%%%%%%%%%%%%%%%%%%%%%%%%%%%
\subsubsection{Depth and inverse depth}

Similar partial differential equations can be derived for the conservation of the depth and inverse depth fields, $\depth(\scoord, t)$ and $\invdepth(\scoord, t)$, respectively. Considering the total derivatives of depth and inverse depth developed in Equations \eqref{eq:totaldev_depth} and \eqref{eq:totaldev_invdepth}, we obtain:
\begin{align}
    \partialdev{\depth}{t} &= -\partialdev{\depth}{\scoord}\tmatrix{\scoord}\hflow + \depth \dotp{\scoord}{\hflow} \\
    \partialdev{\invdepth}{t} &= -\partialdev{\invdepth}{\scoord} \tmatrix{\scoord} \hflow -\invdepth \dotp{\scoord}{\hflow} \label{eq:invdepth_conservation}
\end{align}

The conservation of inverse depth has previously been reported in the literature \cite{2009_Bonnabel,2012_Zarrouati}.

In practice, Equation \eqref{eq:invdepth_conservation} is preferred since $\invdepth$ is a dimensionless ratio  \eqref{eq:inv_depth} that has an intrinsic interpretation as visual measure. 
Note also that \eqref{eq:invdepth_conservation} has the same structure as the first two terms of \eqref{eq:hflow_conservation} which leads to desirable computational properties of the combined system of equations considered in Section \ref{sec:filter_algorithm}.

%%%%%%%%%%%%%%%%%%%%%%%%%%%%%%%%%%%%%%%%%%%%%%%%%%%%%%%%%%%
% FILTER DESIGN
%%%%%%%%%%%%%%%%%%%%%%%%%%%%%%%%%%%%%%%%%%%%%%%%%%%%%%%%%%%

\section{Filter Architecture}
\label{sec:filter_algorithm}

This section introduces a filtering approach for the computation of structure flow in real time using brightness and depth measurements.
We use a pyramidal structure of filter blocks to support large pixel displacements per frame.
Let $H$ denote the number of levels in the pyramid structure indexed as $h = 1, \dots, H$, where $h = 1$ denotes the original resolution level.

The filter state at discrete time index $k$ is denoted by the set
\begin{equation}
    X^k = \left\{ \begin{Bmatrix} \dhflowpyr{k}{1}{} \\ \invdepthpyr{k}{1}{} \end{Bmatrix}, \begin{Bmatrix} \dhflowpyr{k}{2}{} \\ \invdepthpyr{k}{2}{} \end{Bmatrix}, \dots, \begin{Bmatrix} \hflowpyr{k}{H}{} \\ \invdepthpyr{k}{H}{} \end{Bmatrix} \right\}
\end{equation}
Each level $h$ contains an estimate of the inverse depth field $\invdepthpyr{k}{h}{}$ using data of the corresponding pyramid level.
State $\hflowpyr{k}{H}{}$ at top level $H$ represents a coarse estimate of the structure flow based on low resolution down-sampled data.
The flow at each level, $h$, of the pyramid is denoted  $\hflowpyr{k}{h}{}$, however, the flow itself is not used as the dynamic state of the filter for the lower levels of the pyramid.
Instead, the lower level filter states $\dhflowpyr{k}{h}{}$ define the increment to the flow $\hflowpyr{k}{h+1}{}$ given higher resolution data at level $h$ and the the structure flow is reconstructed by
\begin{equation}
    \label{eq:hflow_reconstruction}
    \hflowpyr{k}{h}{} = \hflowpyr{k}{h+1:h}{} + \dhflowpyr{k}{h}{}
\end{equation}
applied recursively for $h = H-1, \dots, 1$.
Here $\hflowpyr{k}{h+1:h}{}$ is the flow at level $h+1$ up-sampled to level $h$.

\begin{figure}[h]
   \begin{subfigure}[b]{0.48\textwidth}
        \centering
        \includegraphics[scale=0.28]{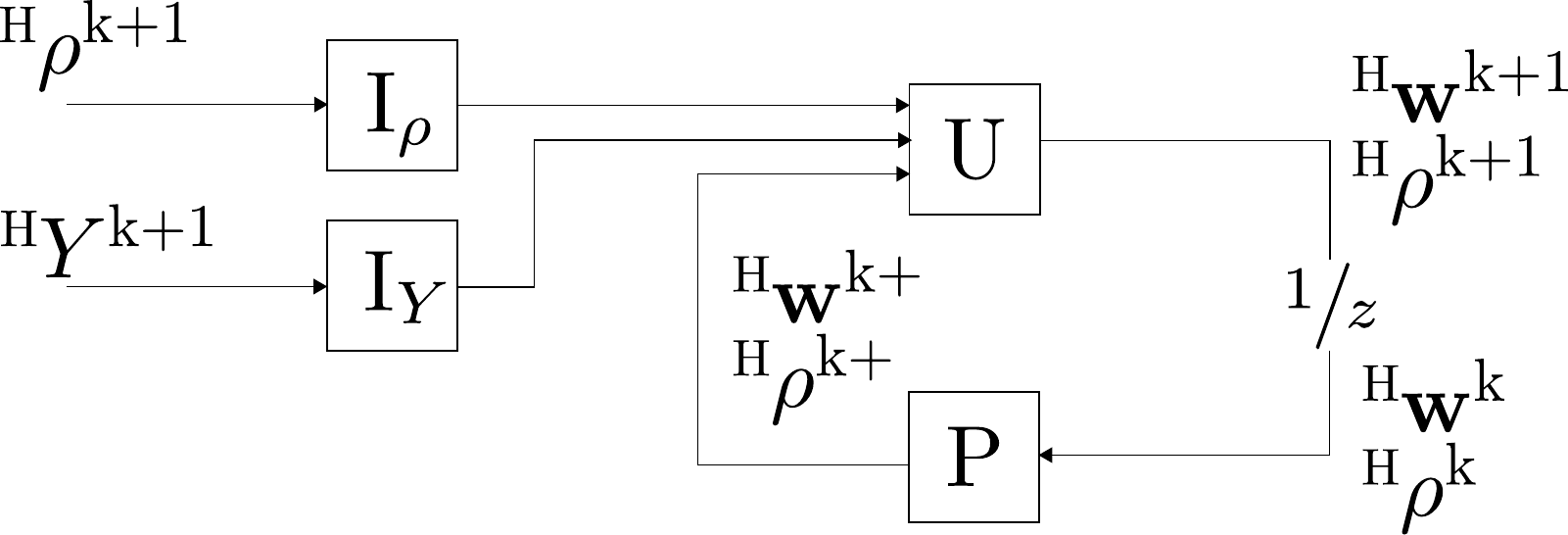}
    \caption{Top level filter loop. The structure flow $\hflowpyr{k}{H}{}$ is used in the prediction stage $[P]$ to compute predictions $\{ \hflowpyr{k+}{H}{}, \invdepthpyr{k+}{H}{} \}$. New raw image and inverse depth measurements are processed at $[I_\img]$ and $[I_\invdepth]$ blocks to extract model parameters. These parameters are used in the update stage $[U]$ to compute new state estimate $\{ \hflowpyr{k+1}{H}{}, \invdepthpyr{k+1}{H}{} \}$.}
    \label{fig:filter_arch_top}
   \end{subfigure}
   \begin{subfigure}[b]{0.48\textwidth}
        \centering
        \includegraphics[scale=0.28]{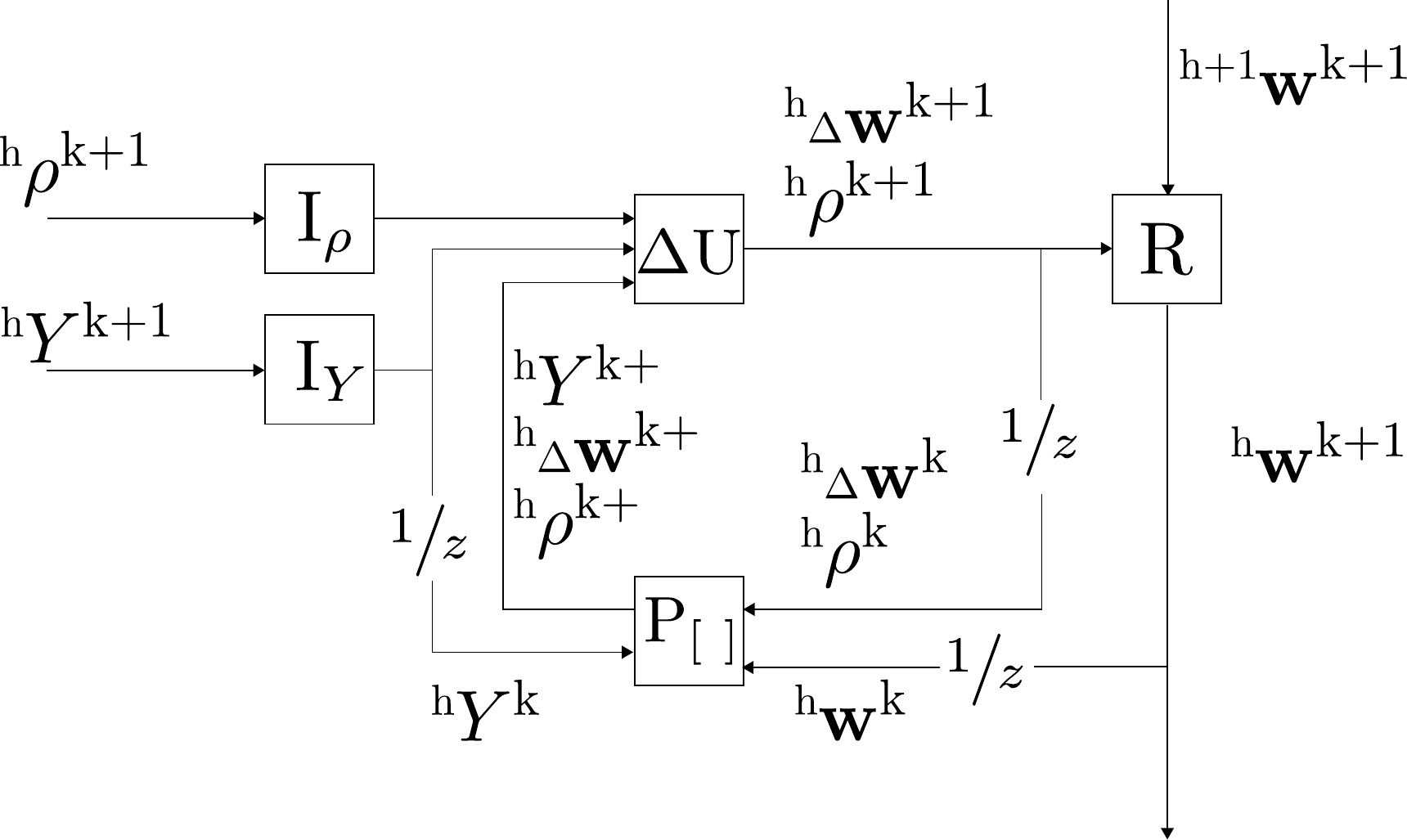}
    \caption{Lower levels filter loop. The reconstruction block $[R]$ reconstructs the flow at this level using the flow from level $h+1$ and state $\dhflowpyr{k+1}{h}{}$. The reconstructed flow is used in the prediction stage $[P_{[]}]$ to create predictions $\imgpyr{k+}{h}{}, \{\dhflowpyr{k+}{h}{}, \invdepthpyr{k+}{h}{} \}$ and $\hflowpyr{k+}{h}{}$. These predictions are combined with image and inverse depth model parameters at the update stage $[\Delta U]$ to create a new estimate $\dhflowpyr{k+1}{h}{}$ and $\invdepthpyr{k+1}{h}{}$.}
    \label{fig:filter_arch_low}
   \end{subfigure}
   \caption{Pyramidal filter architecture.}
\end{figure}

Figure \ref{fig:filter_arch_top} illustrates the filter structure for the topmost level $H$.
Blocks $[I_\img]$ and $[I_\invdepth]$ extract linear model parameters from image and inverse depth raw measurement data. Details of these processes are provided in Sections \ref{ssec:img_model} and \ref{ssec:inv_depth_model}.
In the prediction block $[P]$ (Section~\ref{ssec:state_prediction}), the current estimate $\hflowpyr{k}{H}{}$ is used to create a prediction $\{ \hflowpyr{k+}{H}{}, \invdepthpyr{k+}{H}{} \}$ for the flow and inverse depth state at time $k+1$.
We use  notation $k+$ to refer to the state before new measurements at time $k+1$ are incorporated into it.
The update block $[U]$ (Section~\ref{ssec:state_update}) takes the predicted state $\{ \hflowpyr{k+}{H}{}, \invdepthpyr{k+}{H}{} \}$ and brightness and inverse depth parameters at $k+1$ to create a new state estimate $\{ \hflowpyr{k+1}{H}{}, \invdepthpyr{k+1}{H}{} \}$.

Figure \ref{fig:filter_arch_low} illustrates the filter structure for levels $h=1, \dots, H-1$. The structure flow output is computed at the reconstruction block $[R]$ using the output of the update block $[\Delta U]$ and the flow from level $h+1$ (Equation \eqref{eq:hflow_reconstruction}).
In the prediction block $[P_{[]}]$, the reconstructed flow is used to create predictions of flow $\hflowpyr{k+}{h}{}$, image brightness $\imgpyr{k+}{h}{}$ and state $\{\dhflowpyr{k+}{h}{}, \invdepthpyr{k+}{h}{} \}$. These predictions are plugged into the update block $[\Delta U]$ and combined with new measurement parameters from $[I_\img]$ and $[I_\invdepth]$ blocks to compute a new estimate $\{\dhflowpyr{k+1}{h}{}, \invdepthpyr{k+1}{h}{} \}$.

%%%%%%%%%%%%%%%%%%%%%%%%%%%%%
% SPHEREPIX
%%%%%%%%%%%%%%%%%%%%%%%%%%%%%
\subsection{Spherical image data structure}
\label{ssec:spherepix}

Numerical implementation of the different stages of the algorithm are performed directly on the Spherepix data structure \cite{2016_Adarve_spherepix}, a data structure based on spherical image geometry.
This allows us to preserve the original geometry used to derive the conservation equations in Section \ref{section:conservation_equations}, ensures robustness and consistency of the algorithm over wide field of view, and provides data in an ideal format for incorporation into robotic control algorithms.

\begin{figure}[h]
    \centering
    \includegraphics[width=0.4\textwidth]{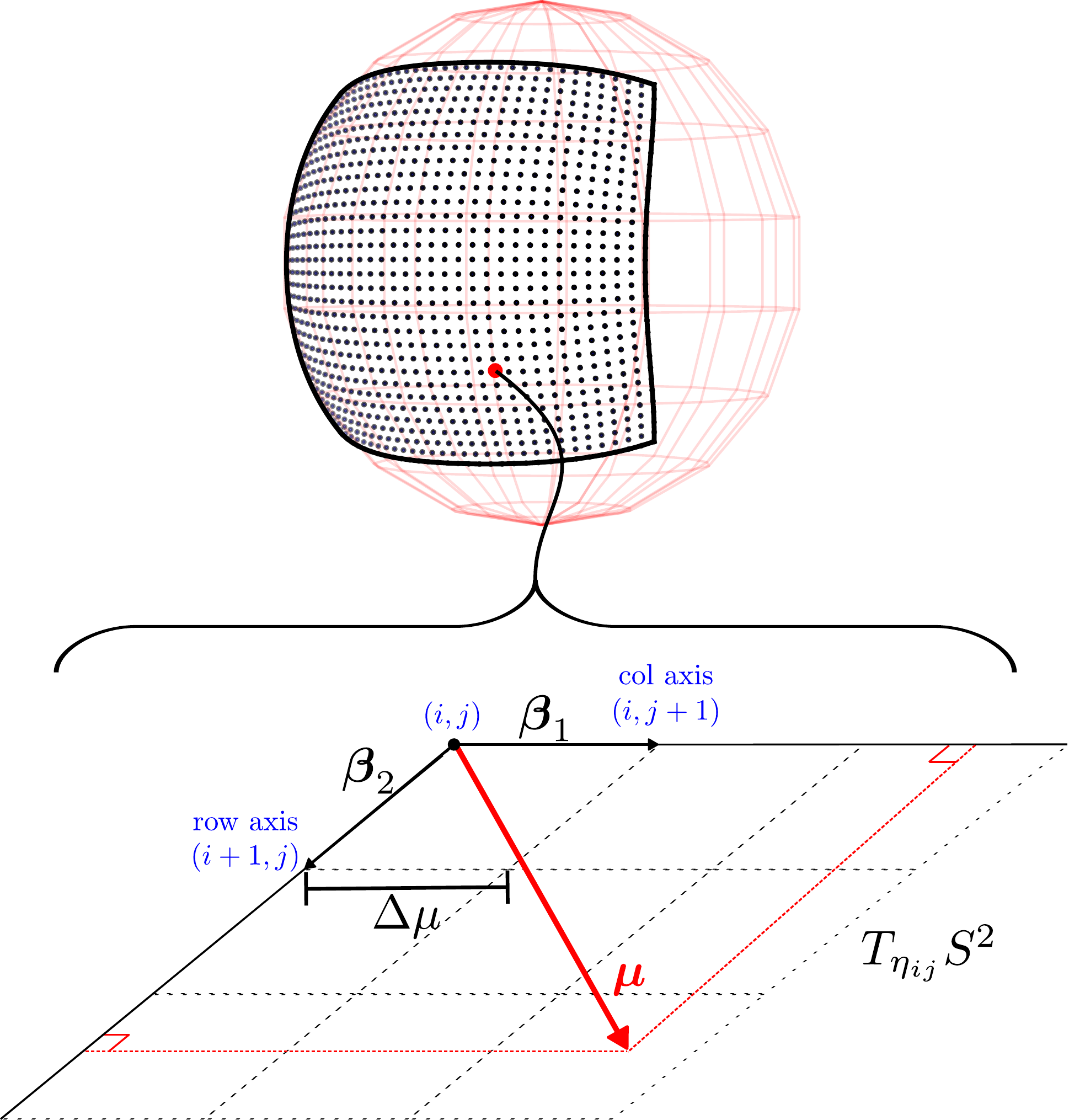}
    \caption{Spherepix grid. Spherical coordinate $\scoord_{ij}$ and its associated tangent plane $\tspace{ij}$. Vector $\tcoord \in \tspace{ij}$ can be represented as a 2-vector $\bcoord$ in the (row, column) direction.}
    \label{fig:spherepix}
\end{figure}

Figure \ref{fig:spherepix} illustrates the relevant properties of a Spherepix image.
At a given pixel $\scoord_{ij} \in \sspace$, neighboring points $\scoord$ are projected onto tangent space coordinates $\tcoord$ using the orthographic projection
\begin{equation}
    \label{eq:tspace_projection}
    \tcoord = \tmatrix{\scoord_{ij}} \scoord.
\end{equation}
As points $\tcoord \in \tspace{ij}$ only have two degrees of freedom, it is convenient to express them as 2D plane coordinates. Let $\bcoord \in \rspace{2}$ be the 2D coordinate representation of $\tcoord$. One has
\begin{equation}
    \bcoord = \betamatrix{\scoord_{ij}}^\top \tcoord
\end{equation}
where $\betamatrix{\scoord_{ij}}$ is an orthonormal basis matrix at $\scoord_{ij}$.
The row entries of $\betamatrix{\scoord_{ij}}$ correspond to the normalized projections of pixels $\scoord_{i,j+1}$ and $\scoord_{i+1,j}$ on the tangent space of $\scoord_{ij}$.
The transformation from 2D to 3D tangent space coordinates is
\begin{equation}
    \label{eq:beta_to_mu}
    \tcoord = \betamatrix{\scoord_{ij}} \bcoord
\end{equation}
since $\betamatrix{\scoord_{ij}}$ is orthonormal and $\tcoord$ lies in the tangent space.
Finally, a scalar $\tcoordsize_{ij} = \vnorm{\tmatrix{\scoord_{ij}} \scoord_{i,j+1} }$ measures the pixel separation between two neighbors in tangent space.

%%%%%%%%%%%%%%%%%%%%%%%%%%%%%
% BRIGHTNESS PARAMETERS
%%%%%%%%%%%%%%%%%%%%%%%%%%%%%
\subsection{Brightness parameters extraction}
\label{ssec:img_model}

Raw image measurements $\img$ from the camera are used to fit a linear brightness model for each pixel $\pixcoord = (i,j)$.
The model parameters correspond to the image gradient $\imggradhat_\pixcoord \in \tspace{\pixcoord}$ and the constant term $\imghat_\pixcoord \in \rspace{}$ at $\pixcoord$.

Computation of these parameters is performed in two stages.
First, the cost function \eqref{eq:img_model} is minimized for parameters $\{ \imggradbcoordhat_\pixcoord, \imghat_\pixcoord\}$ expressed in local 2D plane coordinates.
\begin{equation}
    \label{eq:img_model}
    \varepsilon_\pixcoord = \sum_{q \in \Lambda_{\pixcoord}} G(\pixcoord,q) \| \img_q - \imggradbcoordhat_p (\pixcoord - q) - \hat{\img}_\pixcoord \|^2,
\end{equation}
where $\Lambda_\pixcoord$ is a $5 \times 5$ support window of pixels centered on $\pixcoord$, and $G(\pixcoord, q)$ is a Gaussian weight mask $G = g^\top g$ with $g = [1,4,6,4,1]/16$.
Thanks to the spatial symmetry of the support window and weight function, it is possible to efficiently solve for $\{ \imggradbcoordhat_\pixcoord, \imghat_\pixcoord\}$ as a series of 1D convolutions in the row and column axis.
The two-dimensional image gradient $\imggradbcoordhat_\pixcoord$ is then expressed as a 3D tangent vector $\imggradhat_\pixcoord$ following Equation \eqref{eq:beta_to_mu}
\begin{equation}
    \label{eq:img_gradient}
    \imggradhat_\pixcoord = \tcoordsize_\pixcoord \betamatrix{\pixcoord}^{\top} \imggradbcoordhat_\pixcoord.
\end{equation}

%%%%%%%%%%%%%%%%%%%%%%%%%%%%%
% INVERSE DEPTH PARAMETERS
%%%%%%%%%%%%%%%%%%%%%%%%%%%%%
\subsection{Inverse depth parameters extraction}
\label{ssec:inv_depth_model}

In this paper we assume that depth measurements are available and are inverted for the data fitting process.
We fit a linear model comprised of parameters $\invdepthgradhat_{\pixcoord} \in \tspace{\pixcoord}$ for the inverse depth gradient vector and $\hat{\invdepth}_\pixcoord \in \rspace{}$ for the constant term.
In particular for the computation of $\invdepthgradhat_{\pixcoord}$, one must properly handle object occlusions to remove undesired large gradient vectors.
At each pixel location, we compute the gradient vector on both sides of a discontinuity and select the appropriate one.
That is, the two-dimensional gradient vector $\invDepthBcoordGradHat{}_{\pixcoord} = (\invDepthBcoordGradHat{1}_{\pixcoord}, \invDepthBcoordGradHat{2}_{\pixcoord})^\top$ is computed as follows
\begin{align}
    \invDepthBcoordGradHat{1}_{\pixcoord} &= \left\{ \begin{matrix}
        \diffp{\bcoord_1} \invdepth_\pixcoord  & \text{if}~~ | \diffp{\bcoord_1} \invdepth_\pixcoord | \leq |\diffm{\bcoord_1} \invdepth_\pixcoord| \\
        \diffm{\bcoord_1} \invdepth_\pixcoord & \text{otherwise}
    \end{matrix} \right. \label{eq:dominant_b1} \\
    \invDepthBcoordGradHat{2}_{\pixcoord} &= \left\{ \begin{matrix}
        \diffp{\bcoord_2} \invdepth_\pixcoord  & \text{if}~~ | \diffp{\bcoord_2} \invdepth_\pixcoord | \leq |\diffm{\bcoord_2} \invdepth_\pixcoord| \\
        \diffm{\bcoord_2} \invdepth_\pixcoord & \text{otherwise}
    \end{matrix} \right. \label{eq:dominant_b2}
\end{align}
where $\diffp{}$, $\diffm{}$ denote forward and backward difference operators in the $\bcoord_1$ (column) and $\bcoord_2$ (row) axis. Table \ref{table:diff_operators} defines these as well as the central difference operator.

\begin{table}[H]
\begin{center}
\begin{tabular}{l|ll}
        & \multicolumn{1}{c}{$\delta_{\bcoord_1} u_{ij}$}& \multicolumn{1}{c}{$\delta_{\bcoord_2} u_{ij}$} \\
            \hline
    backward & $\diffm{\bcoord_1} = u_{j} - u_{j-1}$  & $\diffm{\bcoord_2} = u_{i} - u_{i-1}$ \\
    central & $\diffc{\bcoord_1} = u_{j+1} - u_{j-1}$  & $\diffc{\bcoord_2} = u_{i+1} - u_{i-1}$ \\
    forward & $\diffp{\bcoord_1} = u_{j+1} - u_{j}$  & $\diffp{\bcoord_2} = u_{i+1} - u_{i}$
\end{tabular}
\end{center}
\caption{Difference operators in beta coordinates.}
\label{table:diff_operators}
\end{table}
The 3D vector representation $\invdepthgradhat_{\pixcoord} \in \tspace{p}$ is
\begin{equation}
    \label{eq:inv_depth_gradient}
    \invdepthgradhat_{\pixcoord} = \tcoordsize_\pixcoord \betamatrix{\pixcoord}^{\top} \invdepthgradhat_\pixcoord
\end{equation}

The constant parameter term is given by the raw inverse depth data; that is, $\hat{\invdepth}_\pixcoord = \invdepth_\pixcoord$.

%%%%%%%%%%%%%%%%%%%%%%%%%%%%%
% STATE PREDICTION
%%%%%%%%%%%%%%%%%%%%%%%%%%%%%
\subsection{State prediction ($k \rightarrow k+$)}
\label{ssec:state_prediction}

\begin{figure}[!t]
    \centering
    \includegraphics[width=0.6\textwidth]{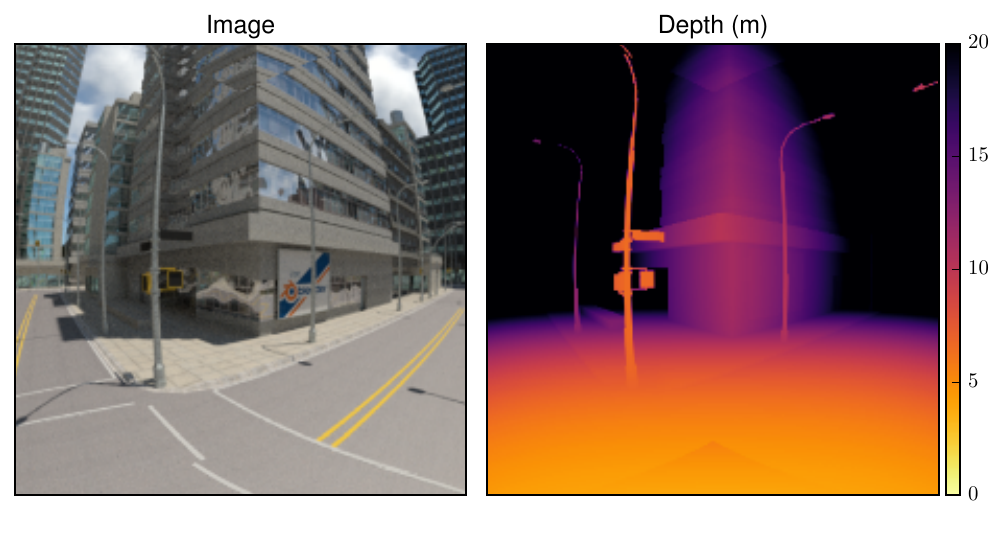} \\
    \includegraphics[width=0.6\textwidth]{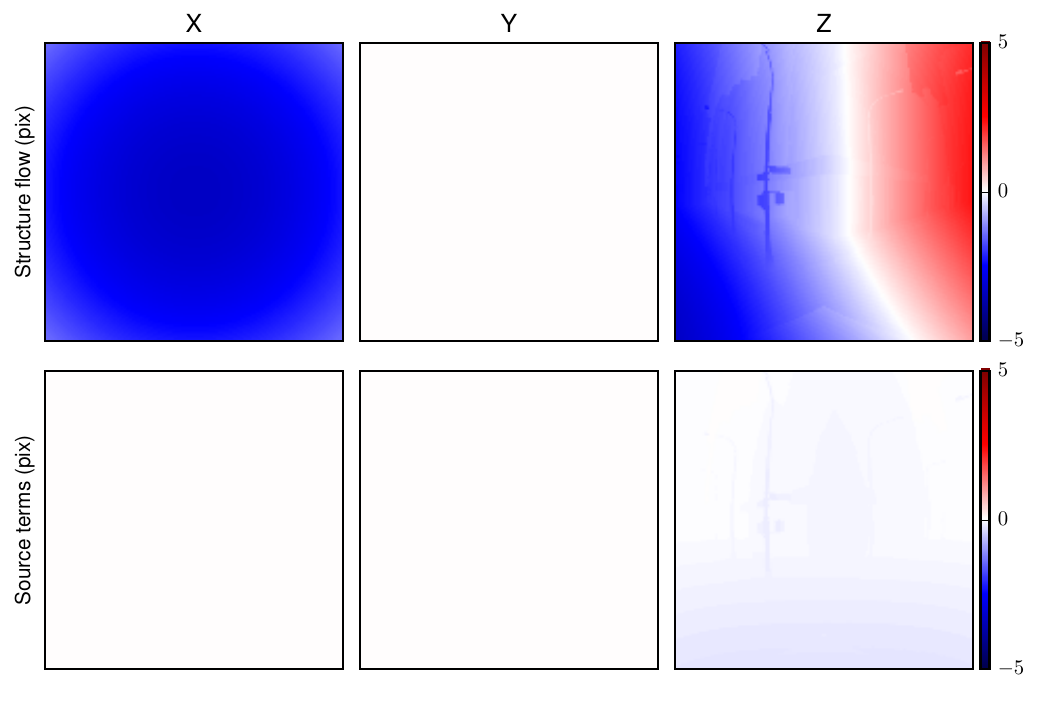}
    \caption{Contribution of the source terms in Equation \eqref{eq:assumption}. The simulated camera runs at 300 Hz and moves with a linear velocity of $5 \sfrac{m}{s}$, acceleration of $1 \sfrac{m}{s^2}$ both in the $z$ direction and angular velocity of $180 \sfrac{\deg}{s}$ in $y$ axis.}
    \label{fig:flow_source_terms}
\end{figure}

The prediction stage of the filter uses the current state estimate to create predictions of the structure flow forward in time.
In the present paper, we choose to ignore the source terms derived from inertial measurements in the PDE \eqref{eq:hflow_propagation_top}.
That is, we assume
\begin{equation}
  \camangvel_\times \hflow - \acc_\hflow \approx 0.
  \label{eq:assumption}
\end{equation}
In practice, for high frame rates and the scenarios that we consider, these terms have sub-pixel magnitudes and are at least an order of magnitude less significant than the other terms in the propagation algorithm.
In particular, the contribution of this term is proportional to the magnitude of $\Omega$ and $\camacc$ and inversely proportional to frame rate.
Thus, for high frame rates and moderate motion these terms become insignificant.
Figure \ref{fig:flow_source_terms} plots the $xyz$ components of Equation \eqref{eq:assumption} for a structure flow field computed from realistic velocities and acceleration values to provide an indication of the validity of this assumption.

For the top level $H$, the prediction block uses the current coarse estimate of structure flow and the inverse depth field as initial conditions $\hflowpyr{}{H}{}(0):= \hflowpyr{k}{H}{}$ and $\invdepthpyr{}{H}{}(0):= \invdepthpyr{k}{H}{}$ for the PDEs \eqref{eq:hflow_conservation} and \eqref{eq:invdepth_conservation} with assumption \eqref{eq:assumption}.
We numerically solve these PDEs to create the state predictions $\hflowpyr{k+}{H}{} := \hflowpyr{}{H}{}(1)$ and $\invdepthpyr{k+}{H}{} := \invdepthpyr{}{H}{}(1)$ for the next frame $k+1$.
The explicit PDEs that we solve at level $H$ are
\begin{align}
    \partialdev{\hflowpyr{}{H}{}}{t} &= -\partialdev{\hflowpyr{}{H}{}}{\scoord} \tmatrix{\scoord}\hflowpyr{}{H}{} -\hflowpyr{}{H}{} \dotp{\scoord}{\hflowpyr{}{H}{}} \label{eq:hflow_propagation_top} \\
    \partialdev{\invdepthpyr{}{H}{}}{t} &= -\partialdev{\invdepthpyr{}{H}{}}{\scoord} \tmatrix{\scoord} \hflowpyr{}{H}{} -\invdepthpyr{}{H}{} \dotp{\scoord}{\hflowpyr{}{H}{}} \label{eq:invdepth_propagation_top}
\end{align}

For lower levels $h=1, \dots, H-1$, the propagation scheme is defined by a PDE system modeling the transport of the structure flow, image brightness and state $\{\dhflowpyr{}{h}{}, \invdepthpyr{}{h}{}\}$ by the reconstructed structure flow $\hflowpyr{}{h}{}$ at each level.
\begin{align}
    \partialdev{\hflowpyr{}{h}{}}{t}  &= -\partialdev{\hflowpyr{}{h}{}}{\scoord} \tmatrix{\scoord}\hflowpyr{}{h}{} -\hflowpyr{}{h}{} \dotp{\scoord}{\hflowpyr{}{h}{}} \label{eq:hflow_propagation_low} \\
    \partialdev{\dhflowpyr{}{h}{}}{t} &= -\partialdev{\dhflowpyr{}{h}{}}{\scoord} \tmatrix{\scoord}\hflowpyr{}{h}{} -\dhflowpyr{}{h}{} \dotp{\scoord}{\hflowpyr{}{h}{}} \label{eq:dhflow_propagation_low} \\
    \partialdev{\invdepthpyr{}{h}{}}{t} &= -\partialdev{\invdepthpyr{}{h}{}}{\scoord} \tmatrix{\scoord} \hflowpyr{}{H}{} -\invdepthpyr{}{h}{} \dotp{\scoord}{\hflowpyr{}{h}{}} \label{eq:invdepth_propagation_low} \\
    \partialdev{\imgpyr{}{h}{}}{t} &= -\partialdev{\imgpyr{}{h}{}}{\scoord} \tmatrix{\scoord}\hflowpyr{}{h}{} \label{eq:img_propagation_low}
\end{align}
Initial conditions are set to $\hflowpyr{}{h}{}(0) := \hflowpyr{k}{h}{}$, $\dhflowpyr{}{h}{}(0) := \dhflowpyr{k}{h}{}$, $\invdepthpyr{}{h}{}(0) := \invdepthpyr{k}{h}{}$ and $\imgpyr{}{h}{}(0) := \imgpyr{k}{h}{}$ and the system is run for one time step unit.
Details on the numerical solution to these equations are provided in Section \ref{ssec:numerical_prediction}.

%%%%%%%%%%%%%%%%%%%%%%%%%%%%%
% STATE UPDATE
%%%%%%%%%%%%%%%%%%%%%%%%%%%%%
\subsection{State update ($k+ \rightarrow k+1$)}
\label{ssec:state_update}

In the update stage of the filter, predictions made for time $k+1$ are corrected using new image and inverse depth measurements. There are two update schemes used; one for top level $H$ to create the updated state $\hflowpyr{k+1}{H}{}$ and one applied to the remaining lower levels to update $\dhflowpyr{K+1}{h}{}$. The update of the inverse depth state $\invdepthpyr{k+1}{h}{}$ follows the same scheme for all pyramid levels.

For level $H$, the update stage creates a new estimate $\hflowpyr{K+1}{H}{}$ using low resolution data available at level $H$.
Cost function \eqref{eq:cost_top} uses low resolution image and inverse depth measurements, as well as the predicted flow from the propagation stage to create a new structure flow estimate $\hflowpyr{k+1}{H}{}$. The cost function
\begin{equation}
    \label{eq:cost_top}
    ^H \varepsilon_\hflow = \gamma_1 \| ^H E_\img \|^2 + \gamma_2 \| ^H E_\invdepth \|^2 + \gamma_3 \| ^H E_t \|^2
\end{equation}
consists of three data terms: $^H E_\img$ (image brightness conservation), $^H E_\invdepth$ (depth measurements) and $^H E_t$ (regularisation based on previous estimate).
Scalar gains $\gamma_1$, $\gamma_2$, $\gamma_3$ control the relative weight of each term.

The image data term $E_\img$ is based on the image conservation PDE \eqref{eq:image_conservation}. It is defined as
\begin{equation}
    \label{eq:img_cost_top}
    ^H E_\img = \imggradhat^{k+1} \tmatrix{\scoord}  \hflow^{k+1} + \tcoordsize^2(\imghat^{k+1} - \imghat^{k})
\end{equation}
where $\imggradhat^{k+1}$, $\imghat^{k+1}$ and $\imghat^{k}$ are the image gradient and constant terms computed at the image model block. The temporal difference term is multiplied by $\tcoordsize^2$ to compensate for the transformation of gradient vector $\imggradhat^{k+1}$ to tangent space coordinates.
This cost term only constrains one degree of freedom of $\hflow$, corresponding to the tangential component of structure flow in the direction of the image gradient vector $\imggradhat$.
This is the well known aperture problem in optical flow estimation.

The inverse depth data term $E_\invdepth$ uses the inverse depth conservation PDE \eqref{eq:invdepth_conservation} to recover the structure flow from state $\invdepth^k$ and measurements and $\invdepthhat^{k+1}$. The data term is
\begin{align}
    ^H E_\invdepth =~&\invdepthgradhat^{k+1} \tmatrix{\scoord}  \hflow^{k+1}  \nonumber \\ &+ \tcoordsize^2\left((\invdepthhat^{k+1} - \invdepth^{k}) + \invdepthhat^{k+1} \dotp{\scoord}{ \hflow^{k+1}} \right) \label{eq:invdepth_cost_top}
\end{align}
Analogously to the image cost term, the scalar component of Equation \eqref{eq:invdepth_cost_top} is multiplied by $\tcoordsize^2$ to compensate for the transformation of vector quantities to 3D tangent space coordinates. As PDE \eqref{eq:invdepth_conservation} includes the effect of $\hflow$ along normal direction $\scoord$, it is possible to recover both tangent and normal components of the flow field.
For the tangent component, it still suffers from a form of aperture problem where only flow in the direction of the inverse depth gradient can be recovered \cite{2002_Spies}.

Finally, the temporal smoothing or regularisation term, $E_t$, includes the predicted flow $\hflowpyr{k+}{H}{}$ as prior solution to the cost function
\begin{equation}
   ^H E_t =  \hflow^{k+1} + \hflow^{k+}
\end{equation}
This regularization term guarantees there is a solution for the structure flow at each pixel. In regions where it is not possible to recover flow from the data, the predicted flow $\hflowpyr{k+}{H}{}$ will act as best estimate for time $k+1$.

Cost function \eqref{eq:LS_update} describes a linear least squares problem with respect to $\hflowpyr{k+1}{H}{}$.
Solving for the minimum of \eqref{eq:cost_top} can be expressed as a $3 \times 3$ linear system
\begin{equation}
    \label{eq:LS_update}
    A \hflowpyr{k+1}{H}{} = b
\end{equation}

After solving Equation \eqref{eq:LS_update}, we apply an average smoothing filter of size $5 \times 5$ to the resulting flow field $\hflowpyr{k+1}{H}{}$ in order to accelerate the diffusion of flow estimates to textureless regions of the image.

For levels $h = 0, \dots, H-1$, the update stage computes new state estimates $\dhflowpyr{k+1}{h}{}$ provided with the predicted state and new measurement data. Cost function \eqref{eq:cost_bottom} is defined as
\begin{equation}
    \label{eq:cost_bottom}
    ^h \varepsilon_\hflow = \gamma_1 \| ^h E_\img \|^2 + \gamma_2 \| ^h E_\invdepth \|^2 + \gamma_3 \| ^h E_t \|^2
\end{equation}
where the data terms for image, inverse depth and temporal smoothing defined are
\begin{align}
    ^h E_\img = ~& \imggradhat^{k+1} \tmatrix{\scoord}  \dhflowpyr{k+1}{h}{} + \tcoordsize^2(\imghat^{k+1} - \imghat^{k+}) \\
    ^h E_\invdepth =~& \invdepthgradhat^{k+1} \tmatrix{\scoord} \dhflowpyr{k+1}{h}{} \nonumber \\
    & + \tcoordsize^2[(\invdepthhat^{k+1} - \invdepth^{k+}) + \invdepthhat^{k+1} \dotp{\scoord}{\dhflowpyr{k+1}{h}{}} ] \\
    ^h E_t = ~& \dhflowpyr{k+1}{h}{} - \dhflowpyr{k+}{h}{}
\end{align}

Equation \eqref{eq:cost_bottom} describes a linear least squares problem with respect to unknown $\dhflowpyr{k+1}{h}{}$, and can be solved following the same procedure used for top level.
Once $\dhflowpyr{k+1}{h}{}$ has been computed, the structure flow at level $h$ is reconstructed using Equation \eqref{eq:hflow_reconstruction}.
Once completed, the resulting flow is cascaded to the next level below until level $h = 1$ is reached and the flow at the original resolution is computed.

The formulation of the inverse depth state update is equal for all pyramid levels.
The proposed cost function
\begin{equation}
    \label{eq:cost_invdepth}
    ^h \varepsilon_\invdepth = \gamma_4\| \invdepthpyr{k+1}{h}{} - \invdepthpyrhat{k+1}{h}{} \|^2 + \gamma_5\| \invdepthpyr{k+1}{h}{} - \invdepthpyr{k+}{h}{} \|^2
\end{equation}
considers both current measurements and the predicted inverse depth from previous frame for estimating new state $\invdepthpyr{k+1}{h}{}$.
Here $\gamma_4$ and $\gamma_5$ are the relative weights between measurements and prediction.
The least squares solution of $\invdepthpyr{k+1}{h}{}$ is
\begin{equation}
    \invdepthpyr{k+1}{h}{} = \frac{\gamma_4 \invdepthpyrhat{k+1}{h}{} + \gamma_5 \invdepthpyr{k+}{h}{}}{\gamma_4 + \gamma_5}
\end{equation}
In places where no measurements are available, $\gamma_4 = 0$ and the updated inverse depth is $\invdepthpyr{k+}{h}{} = \invdepthpyr{k+1}{h}{}$.

%%%%%%%%%%%%%%%%%%%%%%%%%%%%%
% NUMERICAL PROPAGATION
%%%%%%%%%%%%%%%%%%%%%%%%%%%%%
\subsection{Numerical prediction}
\label{ssec:numerical_prediction}

Considering the real-time nature of the structure flow filtering algorithm, we desire a numerical scheme that is fast and robust to noisy input for solving the propagation equations in Section \ref{ssec:state_prediction}.
We propose a numerical scheme based on upwind finite differences \cite{1995_Thomas} following a similar approach to our previous work on optical flow algorithms \cite{2016_Adarve}.

Consider Equation \eqref{eq:hflow_propagation_discrete} as a discrete version of prediction Equations \eqref{eq:hflow_propagation_top} and \eqref{eq:hflow_propagation_low}.
\begin{equation}
    \label{eq:hflow_propagation_discrete}
    \frac{\hflow^{n+1} - \hflow^n}{\Delta t} = -\partialdev{\hflow^n}{\scoord} \tmatrix{\scoord} \hflow^n -\hflow^n \dotp{\scoord}{\hflow^n}
\end{equation}
where $\hflow^{n+1}$ is the unknown flow at time index $n+1$. For convenience in the notation, we omit pyramid level index $h$ and pixel coordinate $(i,j)$, as the scheme runs identically at every level and for each pixel. Index $n$ refers to an internal time index for the numerical scheme running for $n = 0, \dots, N-1$ with $\Delta t = \sfrac{1}{N}$ and $N$ being a parameter describing the number of iterations. Initial conditions of the problem are set to $\hflow^0 := \hflow^k$ and the output of the scheme is $\hflow^{N-1} =: \hflow^{k+}$.

Equation \eqref{eq:hflow_propagation_discrete} is a non-linear PDE on $\hflow$ for which direct application of finite difference methods are not reliable \cite[p. 140]{1999_Thomas}.
To break this non-linearity we use the idea of \emph{dominant optical flow}.
Optical flow in 2D tangent plane coordinates and pixel units is computed as
\begin{equation}
    \label{eq:oflow_numeric}
    \Phi_{ij} := \begin{pmatrix}u_{ij} \\ v_{ij} \end{pmatrix} = \frac{1}{\tcoordsize} \betamatrix{\scoord_{ij}} \tmatrix{\scoord_{ij}} \hflow_{ij}
\end{equation}
The dominant optical flow $\hat{\Phi}_{ij} = (\hat{u}^n_{ij}, \hat{v}^n_{ij})$ is calculated as the largest flow component in the row and column axis
\begin{align}
    \hat{u}_{ij} &= \left\{\begin{matrix}u_{i,j-1} & \text{if}~~ \diffc{\bcoord_1}|u_{ij}| > 0 \\
    u_{i,j+1} & \text{otherwise} \end{matrix} \right. \\
    \hat{v}_{ij} &= \left\{\begin{matrix}v_{i-1,j} & \text{if}~~ \diffc{\bcoord_2}|v_{ij}| > 0 \\
    v_{i+1,j} & \text{otherwise} \end{matrix} \right.
\end{align}
with $\diffc{\bcoord_1}$ and $\diffc{\bcoord_2}$ the central difference operators defined in Table \ref{table:diff_operators}.

Functions $\diff{\bcoord_1}(\cdot; \hat{u})$ and $\diff{\bcoord_2}(\cdot; \hat{v})$ define the upwind finite difference operators in column and row axis. They are computed as
\begin{align}
    \diff{\bcoord_1}(f; \hat{u}) &= \left\{ \begin{matrix} \diffm{\bcoord_1} f & \text{if}~~ \hat{u} > 0 \\
     \diffp{\bcoord_1} f & \text{otherwise} \end{matrix}\right. \\
     \diff{\bcoord_2}(f; \hat{v}) &= \left\{ \begin{matrix} \diffm{\bcoord_2} f & \text{if}~~ \hat{v} > 0 \\
     \diffp{\bcoord_2} f & \text{otherwise} \end{matrix}\right.
\end{align}
These operators evaluate the direction of the dominant flow in the column and row axis and select the adequate upwind difference operator.
By using the dominant optic flow the remaining PDE evolution is linear and has a stable solution.

The iterations of the numerical scheme are as follows.
First, the dominant optical flow $\hat{u}^n$ is calculated from $\hflow^{n}$.  Then, the propagation takes place in the $\bcoord_1$ (column) axis for all points in the spherical grid as follows
\begin{align}
    \hflow^{*} &= \hflow^n - \Delta t \left[ \hat{u}^{n} \diff{\bcoord_1}(\hflow^{n}; \hat{u}^{n}) + \hflow^{n}\dotp{\scoord}{\hflow^{n}} \right] \\
    \dhflow^{*} &= \dhflow^n - \Delta t \left[ \hat{u}^{n} \diff{\bcoord_1}(\dhflow^{n}; \hat{u}^{n}) + \dhflow^{n}\dotp{\scoord}{\hflow^{n}} \right] \\
    \invdepth^{*} &= \invdepth^n - \Delta t \left[ \hat{u}^{n} \diff{\bcoord_1}(\invdepth^{n}; \hat{u}^{n}) + \invdepth^{n}\dotp{\scoord}{\hflow^{n}} \right] \\
    \img^{*} &= \img^n - \Delta t \hat{u}^{n} \diff{\bcoord_1}(\img^{n}; \hat{u}^{n})
\end{align}
producing the propagated fields $\hflow^{*}$, $\dhflow^{*}$, $\invdepth^{*}$ and $\img^{*}$.
Next, the dominant flow $\hat{v}^*$ is calculated from $\hflow^{*}$ and the propagation takes place in the $\bcoord_2$ (row) axis
\begin{align}
    \hflow^{n+1} &= \hflow^{*} - \Delta t \left[ \hat{v}^{*} \diff{\bcoord_2}(\hflow^{*}; \hat{v}^{*}) + \hflow^{*}\dotp{\scoord}{\hflow^{*}} \right] \\
    \dhflow^{n+1} &= \dhflow^{*} - \Delta t \left[ \hat{v}^{*} \diff{\bcoord_2}(\dhflow^{*}; \hat{v}^{*}) + \dhflow^{*}\dotp{\scoord}{\hflow^{*}} \right] \\
    \invdepth^{n+1} &= \invdepth^{*} - \Delta t \left[ \hat{v}^{*} \diff{\bcoord_1}(\invdepth^{*}; \hat{u}^{*}) + \invdepth^{*}\dotp{\scoord}{\hflow^{*}} \right] \\
    \img^{n+1} &= \img^{*} - \Delta t \hat{v}^{*} \diff{\bcoord_2}(\img^{*}; \hat{v}^{*})
\end{align}
The numerical scheme is stable if inequality in Equation \eqref{eq:numerical_stability} is satisfied for all points in the grid for all $n$.
\begin{equation}
    \label{eq:numerical_stability}
    \Delta t \max_{n,i,j} \left\{ |\hat{u}^{n}_{ij}|, |\hat{v}^{n}_{ij}| \right\} \le 1
\end{equation}
Given this stability condition where $\Delta t = \sfrac{1}{N}$, one needs to choose an suitable number of iterations $N$ of the numerical scheme depending on the maximum flow expected in the application.
In particular, as the camera frame rate increases, the magnitude of the estimated flow decreases and hence, less numerical iterations are required, leading to excellent numerical properties for real-time high-speed vision applications.

%%%%%%%%%%%%%%%%%%%%%%%%%%%%%%%%%%%%%%%%%%%%%%%%%%%%%%%%%%%
% EXPERIMENTAL VALIDATION
%%%%%%%%%%%%%%%%%%%%%%%%%%%%%%%%%%%%%%%%%%%%%%%%%%%%%%%%%%%
\section{Experimental Validation}
\label{sec:experimental_validation}

\subsection{GPU implementation}
We developed a GPU implementation of the filtering algorithm for its deployment in real-time applications.
Computations within each stage of the algorithm are pixel independent and can easily be parallelized on GPU hardware.
The different stages of the algorithm are programmed and connected using a pipeline pattern.
We run the implementation on a Desktop computer with an Intel i7-4790K CPU and a Nvidia GTX-780 graphics card with 2304 CUDA cores and 3GB of memory.

\subsection{Ground truth evaluation}

\begin{figure}[h]
   \begin{subfigure}[b]{1.0\textwidth}
        \centering
        \includegraphics[scale=0.5]{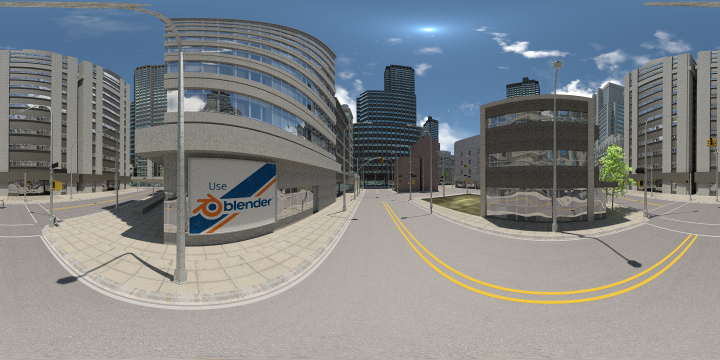}
    \caption{Panoramic rendering.}
    \label{fig:urban_canyon_panoramic}
   \end{subfigure}

   \begin{subfigure}[b]{1.0\textwidth}
        \centering
        \includegraphics[scale=0.6]{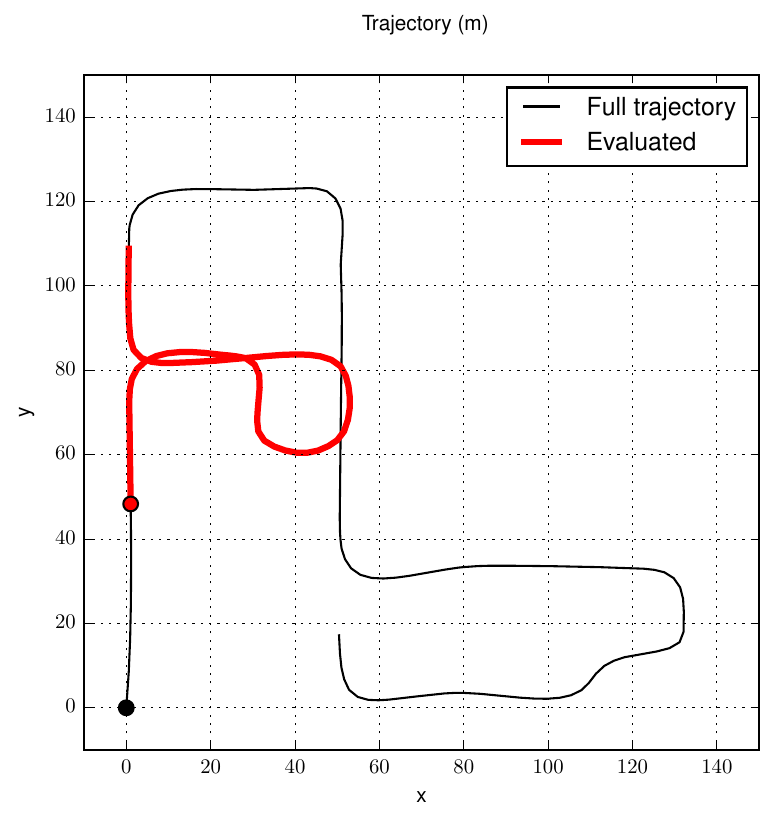}
    \caption{Vehicle trajectory.}
    \label{fig:urban_canyon_trajectory}
   \end{subfigure}
   \caption{Urban Canyon dataset.}
\end{figure}

To validate the accuracy of the proposed algorithm, we require video sequences recorded at high frame rates that sample the dynamics of the environment smoothly.
We used the Blender source files of the Urban Canyon Dataset from Zhang \textit{et.~al.} \cite{2016_Zhang} to render a photo-realistic city environment in which a 300Hz camera is attached to the front part of a moving vehicle. For the evaluation, we consider a subset of the trajectory (Figure \ref{fig:urban_canyon_trajectory}) composed of 10000 frames.
Images are rendered using a 360 degrees panoramic camera and then mapped to a front facing spherepix grid of $512 \times 512$ pixels.

The structure flow ground truth $\hflow_{\text{gt}}$ is computed for each pixel from the rendered depth map and the camera linear and angular velocity using Equation \eqref{eq:homogeneous_flow} with $\pvel \equiv 0$.

We use \emph{Root Mean Square} error (RMSE) and \emph{Absolute Angular Error} (AAE) to measure the accuracy of our algorithm relative to ground truth. The RMSE is calculated by
\begin{equation}
     \label{eq:RMSE_vel}
    \text{RMSE} = \vnorm{\frac{ \hflow_{\text{gt}} - \hflow}{\tcoordsize}}
\end{equation}
and the AAE is computed as
\begin{equation}
    \text{AAE} = \arccos\left( \frac{1 + \dotp{\hflow_{\text{gt}}}{\hflow}}{\sqrt{1 + \hflow_{\text{gt}}} \sqrt{1 + \hflow} } \right)
\end{equation}

For displaying purposes, we split the 3D structure flow $\hflow$ into tangent and normal components.
Tangent flow $\tmatrix{\scoord} \hflow(\scoord,t)$, that is, the structure flow projected onto the tangent space, is transformed to 2D plane coordinates and is expressed in pixel units
\begin{equation}
    \label{eq:tangent_flow}
    \hflow_\perp = \frac{\betamatrix{\scoord}^\top \tmatrix{\scoord}\hflow} {\tcoordsize}.
\end{equation}
It is possible to use the standard optical flow color wheel encoding \cite{2011_Baker} to represent the tangent flow vector field as a color image.
The normal flow, Equation \eqref{eq:normal_flow}, is the component of the structure flow along direction $\scoord$
\begin{equation}
    \hflow_\| = \frac{\dotp{\scoord}{\hflow}} {\tcoordsize} \label{eq:normal_flow}
\end{equation}
It is also measured in pixel units and plotted as a scalar field.

\begin{figure*}[t]
    \centering
    \includegraphics[width=0.97\textwidth]{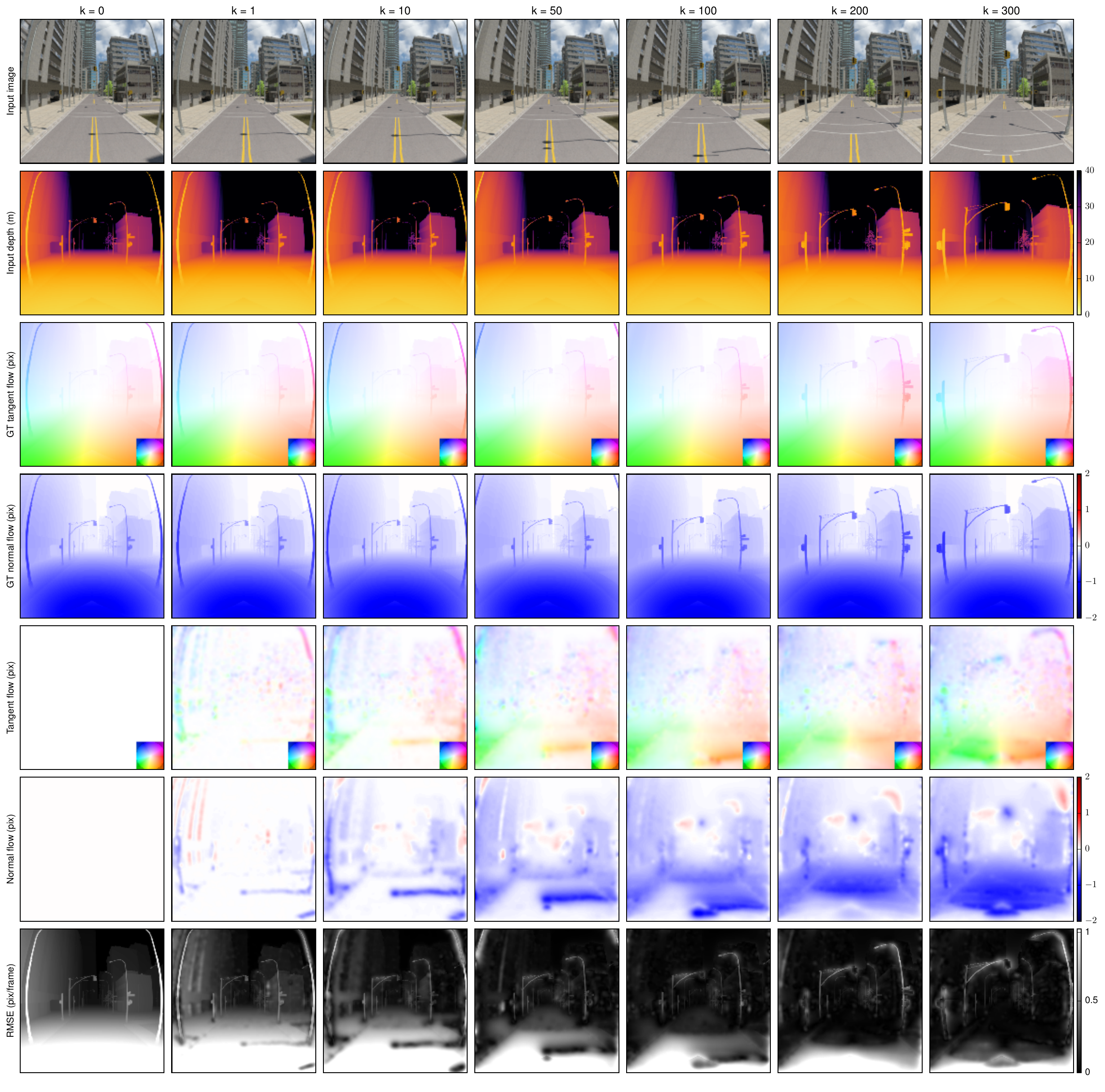}
    \caption{Results on the Urban Canyon dataset. The sequence plots the estimated structure flow and ground truth for the first second (300 frames) of video.}
    \label{fig:results_urban_canyon}
\end{figure*}

Figure \ref{fig:results_urban_canyon} shows a selection of the first 300 images corresponding to 1 second of real-life time.
The initial condition of the structure flow is set to zero.
The flow is first identified at regions with brightness or depth discontinuities and it is diffused to textureless regions.
Over time, a dense flow field is estimated and maintained using new measurements.
Convergence of the algorithm can be visualized through the $\text{RMSE}$ field plot (bottom row).
Regions with higher error are localized at occlusion boundaries, where the algorithm needs to re-identify flow.

\begin{figure}[h]
   \begin{subfigure}[b]{1.0\textwidth}
	    \centering
        \includegraphics[scale=0.6]{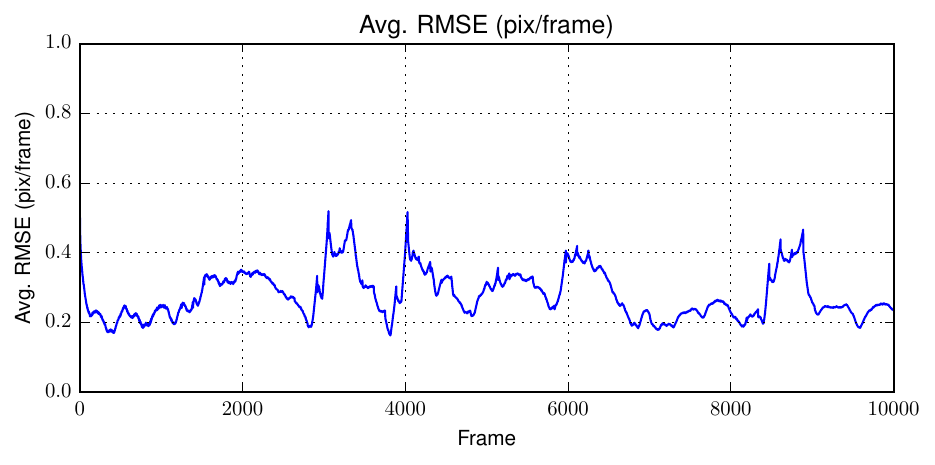}
        \caption{Average Root Mean Square Error.}
        \label{fig:results_urban_canyon_RMSE}
   \end{subfigure}

   \begin{subfigure}[b]{1.0\textwidth}
   	 	\centering
        \includegraphics[scale=0.6]{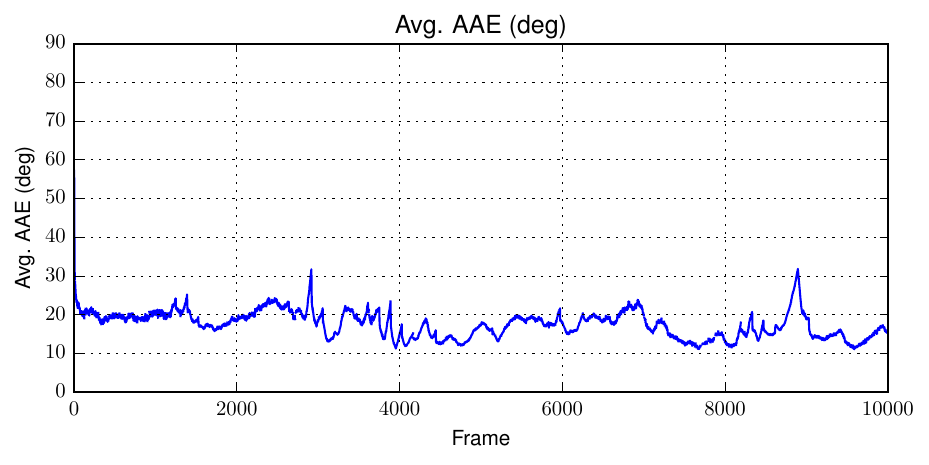}
        \caption{Average Absolute Angular Error.}
        \label{fig:results_urban_canyon_AAE}
   \end{subfigure}
   \caption{Error metrics for the Urban Canyon dataset.}
\end{figure}

Figure \ref{fig:results_urban_canyon_RMSE} plots the average $\text{RMSE}$ error for the evaluated sequence.
After approximately 150 frames, the filter algorithm has converged to a dense flow estimate.
After this, the $\text{RMSE}$ remains stable around 0.3 pixels and increases in parts of the sequence with high camera rotation, as the algorithm needs to adapt to the changes in velocity.
Figure \ref{fig:results_urban_canyon_AAE} displays the average AAE, which stays approximately constant at 20 degrees.

In terms of runtime performance, Tables \ref{table:runtime_pyramid_1} and \ref{table:runtime_pyramid_2} shows the estimated runtime and frame rates of the algorithm configured with one and two pyramid levels respectively. For each configuration, we vary the maximum flow allowed in the algorithm, which affects the number of numerical iterations required in the propagation stage. Runtime is considered as the sum of time required to map image and depth measurements onto the spherepix image plus the runtime of the flow filter itself. We exclude memory transfer operations between CPU and GPU memory spaces, as they can be overlaid with the execution of GPU kernel code.

\begin{table}
\centering
\begin{tabular}{c|cccc}
    Max flow & Runtime & Frequency & Throughput \\
     (pixels) & (ms) & (Hz) & (Mpix/s) \\ \hline
    1 & 0.929 \sfrac{+}{-} 0.3 & 1133.9 \sfrac{+}{-} 176.9 & 283.483 \sfrac{+}{-} 44.2 \\
    2 & 1.029 \sfrac{+}{-} 0.2 & 998.1 \sfrac{+}{-} 111.5 & 249.521 \sfrac{+}{-} 27.9 \\
    4 & 1.341 \sfrac{+}{-} 0.2 & 755.6 \sfrac{+}{-} 67.1 & 188.896 \sfrac{+}{-} 16.8 \\
\end{tabular}
\caption{Runtime performance with one pyramid level at $512 \times 512$ resolution. Smooth iterations set to 2.}
\label{table:runtime_pyramid_1}
\end{table}

\begin{table}
\centering
\begin{tabular}{c|cccc}
    Max flow & Runtime & Frequency & Throughput \\
     (pixels) & (ms) & (Hz) & (Mpix/s) \\ \hline
    1 & 1.355 \sfrac{+}{-} 0.2 & 747.4 \sfrac{+}{-} 65.5 & 186.838 \sfrac{+}{-} 16.4 \\
    2 & 1.405 \sfrac{+}{-} 0.2 & 720.5 \sfrac{+}{-} 63.4 & 180.126 \sfrac{+}{-} 15.9 \\
    4 & 1.501 \sfrac{+}{-} 0.2 & 674.3 \sfrac{+}{-} 59.0 & 168.584 \sfrac{+}{-} 14.7 \\
    8 & 1.700 \sfrac{+}{-} 0.2 & 594.7 \sfrac{+}{-} 51.2 & 148.680 \sfrac{+}{-} 12.8 \\
\end{tabular}
\caption{Runtime performance with two pyramid levels at $512 \times 512$ resolution. Smooth iterations set to [2, 4] for bottom and top level respectively.}
\label{table:runtime_pyramid_2}
\end{table}

On average, the algorithm runs approximately at 600 Hz with 8 pixels maximum flow.
Although the number of operations per pixel is deterministic given parameters such as smooth iterations and maximum flow, there is some variability in the reported runtime, which can be attributed to GPU task swapping and CPU process scheduling.

\subsection{Evaluation on real-life data}

\begin{figure*}[!t]
    \centering
    \includegraphics[width=1.0\textwidth]{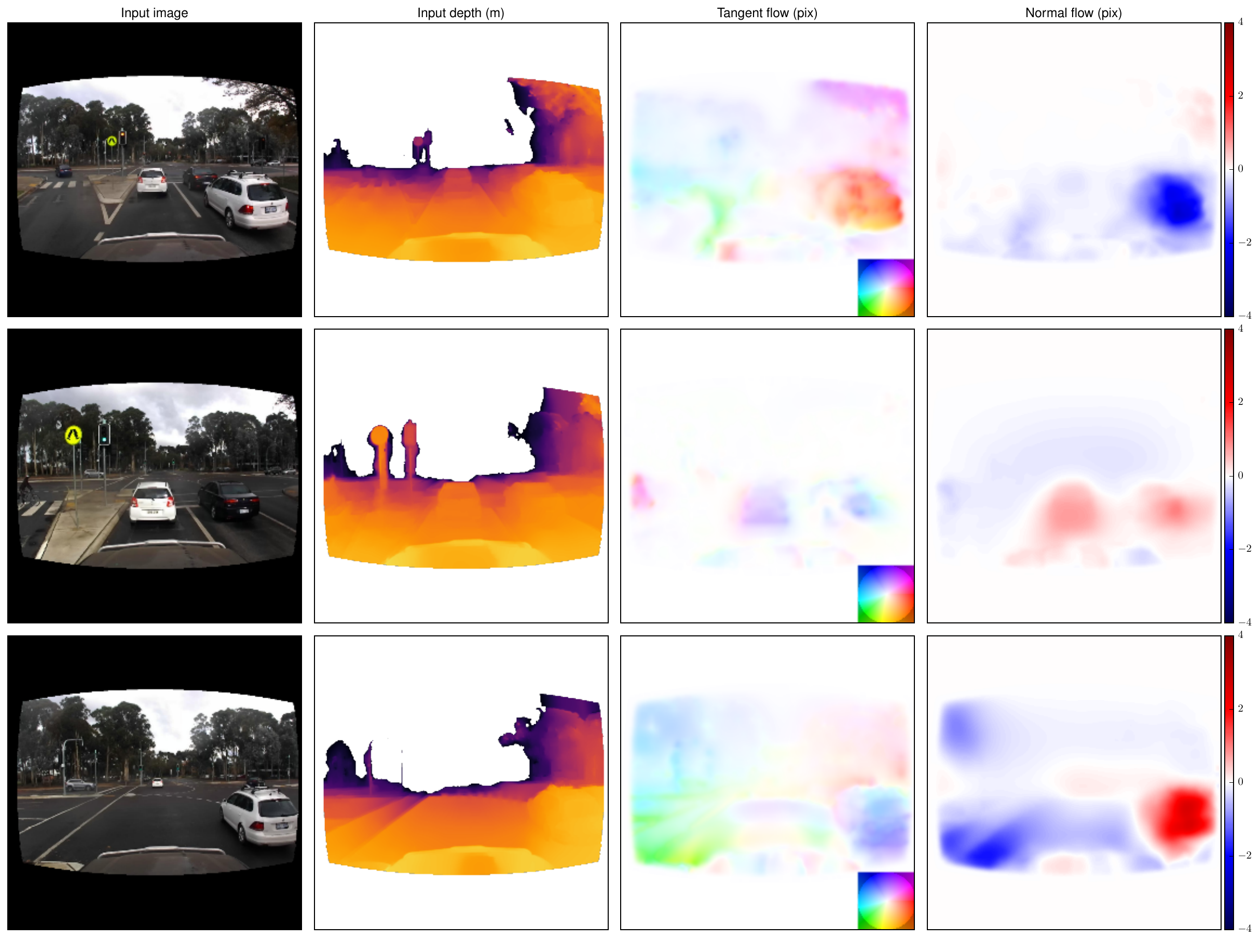}
    \caption{Results on real-life stereo video sequences. The camera captures $1440 \times 720$ images at 60 Hz and computes depth. The brightness and estimated depth are used in the filter algorithm to compute the underlying structure flow. The normal flow component of the structure flow offers immediate information about objects getting close to or far from the camera (rows 1 and 3).}
    \label{fig:results_zed}
\end{figure*}

We performed a set of experiments using real life data captured using a ZED Stereo Camera. Stereo video is captured at a resolution of $1440 \times 720$ per camera at 60 Hz. We used the provided software to extract depth from the stereo video and use it in our structure flow algorithm.
In this case, we use a spherepix patch of $1024 \times 1024$ resolution which approximately matches the resolution and field of view of the input camera. The algorithm is configured with two pyramid levels and maximum flow of 8 pixels. Figure \ref{fig:results_zed} displays the results for some selected scenes in driving scenarios with moving vehicles.

An important difference of the estimated flow compared to conventional optical flow is the ability to directly distinguish objects getting closer or farther from the camera, as can be observed in the normal flow plots (column 4).
Table \ref{table:runtime_pyramid_2_1024} shows the average runtime for this configuration varying the maximum flow. On average, our method achieves frame rates of 172 Hz, which is easily sufficient for use in control loops for even highly dynamic mobile robotic vehicles.

\begin{table}[h]
\centering
\begin{tabular}{c|cccc}
    Max flow & Runtime & Frequency & Throughput \\
     (pixels) & (ms) & (Hz) & (Mpix/s) \\ \hline
    1 & 4.683 \sfrac{+}{-} 0.2 & 213.889 \sfrac{+}{-} 8.2 & 213.889 \sfrac{+}{-} 8.2 \\
    2 & 4.826 \sfrac{+}{-} 0.2 & 207.480 \sfrac{+}{-} 6.9 & 207.480 \sfrac{+}{-} 6.9 \\
    4 & 5.154 \sfrac{+}{-} 0.2 & 194.258 \sfrac{+}{-} 6.3 & 194.258 \sfrac{+}{-} 6.3 \\
    8 & 5.815 \sfrac{+}{-} 0.2 & 172.149 \sfrac{+}{-} 5.1 & 172.149 \sfrac{+}{-} 5.1 \\
\end{tabular}
\caption{Runtime performance with two pyramid levels at $1024 \times 1024$ resolution. Smooth iterations set to [2, 4] for bottom and top level respectively.}
\label{table:runtime_pyramid_2_1024}
\end{table}

\section{Conclusions and Future Work}
\label{sec:conclusions}

This article introduced the structure flow field as the three-dimensional vector field encoding the velocity of the surrounding environment relative to the camera body fixed frame and scaled by the inverse depth.
We presented partial differential equations expressed in spherical geometry modeling the temporal evolution of the structure flow field, image brightness, depth and inverse depth fields.
These PDEs can be used both to derive predictors to propagate these quantities forward in time, or as innovation terms to estimate the structure flow from measured data.

A real-time algorithm for computing structure flow is proposed.
The algorithm is based on a filtering architecture that incrementally computes the flow from temporal sequences of brightness and depth data.
The algorithm achieves frame rates of the order of 600 Hz at $512 \times 512$ pixel resolution and 172 Hz at $1024 \times 1024$ for flow vectors up to 8 pixels.
Experimental validation of the algorithm using simulated high-speed video with ground truth is provided, as well as results on real-life video sequences using a stereo camera.

Assumption \eqref{eq:assumption} is a practical assumption that allows implementation of the algorithm in the absence of a calibrated and time-synchronised inertial measurement unit.
Such a system was unavailable for the field tests that we undertook and would have involved considerable complexity in systems integration.
It is clear that inclusion of the exogenous acceleration terms would improve results, however, we do not believe that the improvement will be particularly significant due to the relatively small value of these terms (Figure \ref{fig:flow_source_terms}).
More significantly, the current formulation of the filter algorithm relies on the availability of depth measurements.
A natural extension is to express the cost functions relying on depth measurements in terms of image disparity for a coupled stereo pair.
Such approach will allow us to formulate the state update equations purely on image data and is the topic of ongoing investigation.

\section{Acknowledgments}

Thanks to Henri Rebecq from the Robotic and Perception Group at University of Zurich for sharing the Blender source files of the Urban Canyon dataset.
This research was supported by the Australian Research Council through the ``Australian Centre of Excellence for Robotic Vision'' CE140100016.

%===============================================================================

%===============================================================================
%% bibliography
%% Use the \bibliographystyle{alpha} to compile the bibligraphy in the main directory.
%% Fetch the .bbl file from the directory above.
%% copy the .bbl file directly in here
\bibliographystyle{alpha}
\bibliography{arxiv_Adarve_Hflow}

%===============================================================================

%%%%%%%%%%%%%%%%%%%%%%%%%%%%%%%%%%%%%%%%%%%%%%%%%%%%%%%%%%%%%%%%%%%%%%%%%%%
\end{document}